\documentclass[journal]{IEEEtran}
\usepackage{times}
\usepackage{epsfig}
\usepackage{graphicx}
\usepackage{amsmath}
\usepackage{amssymb}
\usepackage{subfigure}
\usepackage{color}
\usepackage{multirow}      
\usepackage{multicol} 
\usepackage{booktabs}
\usepackage{hyperref}
\usepackage{threeparttable}
\usepackage[numbers,sort&compress]{natbib}

\newcommand{\wh}[1]{{\color{black}#1}}

\begin{document}
	
	\title{Advancing Image Understanding in Poor Visibility Environments: A Collective Benchmark Study}
	
	\markboth{IEEE Transactions on Image Processing, Draft}%
	{Shell \MakeLowercase{\textit{et al.}}}
	
	\author{Wenhan Yang*, Ye Yuan*\thanks{*The first two authors Wenhan Yang and Ye Yuan contributed equally.
    Ye Yuan and Wenhan Yang helped prepare the dataset proposed for the UG2+ Challenges, and were the main responsible members for UG2+ Challenge 2019 (Track 2) platform setup and technical support.
	Wenqi Ren, Jiaying Liu, Walter J. Scheirer, and Zhangyang Wang were the main organizers of the challenge and helped prepare the dataset, raise sponsors, set up evaluation environment, and improve the technical submission.
	Other authors are the group members of winner teams in UG2+ challenge Track 2 contributing to the winning methods.
	
	Wenhan Yang and Jiaying Liu are with the Wangxuan Institute of Computer Technology, Peking University, Beijing 100080, China (e-mail: yangwenhan@pku.edu.cn; liujiaying@pku.edu.cn).

	Ye Yuan and Zhangyang Wang are with the Department of Computer Science and Engineering,
	Texas A\&M University, TX 77843 USA (e-mail: atlaswang@tamu.edu; ye.yuan@tamu.edu).

	Wenqi Ren is with State Key Laboratory of Information Security, Institute of Information Engineering, Chinese Academy of Sciences (e-mail: rwq.renwenqi@gmail.com).

	W. Scheirer is with the Department of Computer Science and Engineering, University of Notre Dame, Notre Dame, IN, 46556 (e-mail: walter.scheirer@nd.edu).

%
%
%
%
%
%
		},
		Wenqi Ren, Jiaying Liu, Walter J. Scheirer, Zhangyang Wang\thanks{Correspondence should be addressed to: Zhangyang Wang $\langle$\textit{atlaswang @tamu.edu}$\rangle$, and Walter J. Scheirer $\langle$\textit{walter.scheirer@nd.edu}$\rangle$},
		\\
		Taiheng Zhang, Qiaoyong Zhong, Di Xie, Shiliang Pu, 
		Yuqiang Zheng, Yanyun Qu, Yuhong Xie, Liang Chen, Zhonghao Li, and Chen Hong, 
		Hao Jiang, Siyuan Yang, Yan Liu, Xiaochao Qu, Pengfei Wan, 
		Shuai Zheng, Minhui Zhong, Taiyi Su, Lingzhi He, Yandong Guo, Yao Zhao, Zhenfeng Zhu,
		Jinxiu Liang, Jingwen Wang, Tianyi Chen, Yuhui Quan, Yong Xu, 
		Bo Liu, Xin Liu, Qi Sun, Tingyu Lin, Xiaochuan Li, Feng Lu, Lin Gu, 
		Shengdi Zhou, Cong Cao, 
		Shifeng Zhang, Cheng Chi, Chubin Zhuang, Zhen Lei, Stan Z. Li, Shizheng Wang, Ruizhe Liu, Dong Yi, Zheming Zuo, Jianning Chi, Huan Wang, Kai Wang, Yixiu Liu, Xingyu Gao, Zhenyu Chen, Chang Guo, Yongzhou Li,  Huicai Zhong, Jing Huang, Heng Guo, Jianfei Yang, Wenjuan Liao, Jiangang Yang, Liguo Zhou, Mingyue Feng, Likun Qin
	}
	
	\maketitle
	
	\begin{abstract}
		Existing enhancement methods are empirically expected to help the high-level end computer vision task: however, that is observed to not always be the case in practice. 
		We focus on object or face detection in poor visibility enhancements caused by bad weathers (haze, rain) and low light conditions. To provide a more thorough examination and fair comparison, we introduce three benchmark sets collected in real-world hazy, rainy, and low-light conditions, respectively, with annotated objects/faces. We launched the UG$^{2+}$ challenge Track 2 competition in IEEE CVPR 2019, aiming to evoke a comprehensive discussion and exploration about whether and how low-level vision techniques can benefit the high-level automatic visual recognition in various scenarios. To our best knowledge, this is the first and currently largest effort of its kind. Baseline results by cascading existing enhancement and detection models are reported, indicating the highly challenging nature of our new data as well as the large room for further technical innovations. Thanks to a large participation from the research community, we are able to analyze representative team solutions, striving to better identify the strengths and limitations of existing mindsets as well as the future directions.
	\end{abstract}
	
	
	\IEEEpeerreviewmaketitle
	
	\section{Introduction}
	
	The arrival of the big data era brings us mass diverse applications and spawns a series of demands in both human and machine visions.
	On one hand, new applications in consumer electronics~\cite{machine_learning_survey}, 
	such as TV broadcasting, movies, video-on-demand, \textit{etc}., expect continuous efforts to improve the human visual experience.
	On the other hand, many applications of smart cities and Internet of things~(IoT), such as surveillance video, autonomous/assisted driving, unmanned aerial vehicles~(UAVs), \textit{etc.}, call for more effective and stable machine vision-based sensing and understanding~\cite{zhu2018visdrone}.
	As a result, it is a critical issue to explore the general framework that can benefit two kinds of tasks simultaneously and make them mutually beneficial.
	
	However, most of the existing researches still aim to solve the problems in their own routes separately.
	In early researches~\cite{ScSR,BM3D,Lowe04distinctiveimage},
	their models are not capable enough to consider beyond their own purposes (only for one of human vision or machine vision).
	Even in the recent decade, most of the enhancement methods, \textit{e.g.},  dehazing~\cite{DehazeNet,Ren-ECCV-2016,AOD_Net,dark_Channel},
	deraining~\cite{chen2013generalized,luo2015removing,li2016rain,zhang2018density,Chang_2017_ICCV,Fu_2017_CVPR,Fu_TIP,yang2017deep,yang2019tpami,yang2019raindrop},
	illumination enhancement~\cite{Li_2017_SRRM,Ren_2018_SD,yang_2016_dict,LLNet,MSRNet_2017,EnlightenGAN},
	image/video compression~\cite{Hu2020Coarse,Liu2019Gen,Liu2019Interp},
	compression artifact removal~\cite{dmcnn,rethinking},
	and copy-move forgery detection~\cite{forgery_detection}, only target human vision and most of image/video understanding and analytics methods, \textit{e.g.}
	classification~\cite{residual_network},
	segmentation~\cite{deep_lab},
	action recognition~\cite{two_stream},
	and human pose estimation~\cite{Guler_2018_CVPR},
	are considered to take the clean and high-quality images as the input of a system.
	
	When the improved computation power and the new emerging data-driven approaches push for the progress of applying the existing state-of-the-art methods to industrial trials, lack of consideration on human and machine vision jointly leads to the observed fragility of real systems.
	Taking autonomous driving as an example: the industry players have been tackling the challenges posed by inclement weathers; However, heavy rain, haze or snow will still obscure the vision of on-board cameras and create confusing reflections and glare, leaving the state-of-the-art self-driving cars in the struggle~\cite{Forbes1}. 
	Another illustrative example can be found in city surveillance: even the commercialized cameras adopted by governments appear fragile in challenging weather conditions~\cite{wtop1}.
	
	The largely jeopardized performance of visual sensing is caused by two aspects: \textit{inconsistency between training and testing data}, \textit{inappropriate guidance for network training}. 
	First, most current vision systems are designed to perform in clear environments but the real-world scenes are 
	unconstrained and might include dynamic degradation, \textit{e.g.} moving platforms, bad weathers, and poor illumination, which are not covered in the training data for pretrained models.
	Second, the existing data-driven methods largely rely on task-driven tuning and extract task-related features.
	Therefore, they are sensitive to unseen contents and conditions.
	Once the model faces different degradation to the trained one or has a different target, \textit{e.g.} switching from human vision to machine vision, the performance of the pretrained model might degrade much.

	To face the real world in practical applications, a dependable vision system must reckon with the entire spectrum of complex unconstrained outdoor environments, instead of just working in limited scenes. 
	However, at this point, existing academia and industrial solutions see significant gaps from addressing the above-mentioned pressing real-world challenges, and a systematic consideration and collective effort for identifying and resolving those bottlenecks that they commonly face have also been absent. 
	Considering that, it is highly desirable to study to what extent, and in what sense, such challenging visual conditions can be coped with, for the goal of achieving robust visual sensing and understanding in the wild, which benefits security/safety, autonomous driving, robotics, and an even broader range of signal/image processing applications. 
	
	One primary challenge arises from the \textbf{Data} aspect. 
	Those challenging visual conditions usually give rise to nonlinear and data-dependent degradations that will be much more complicated than the well-studied noise or motion blur. The state-of-the-art deep learning methods are typically hungry for training data. The usage of synthetic training data has been prevailing, but may inevitably lead to domain shifts \cite{liu2018improved}. Fortunately, those degradations often follow some parameterized physical models.
	That will naturally motivate a combination of model-based and data-driven approaches. In addition to training, the lack of real-world test sets (and consequently, the usage of potentially oversimplified synthetic sets) has limited the practical scope of the developed algorithms.
	
	The other main challenge is found in the \textbf{Goal} side.
	Most restoration or enhancement methods cast the handling of those challenging conditions as a post-processing step of signal restoration or enhancement after sensing, and then feed the restored data for visual understanding. The performance of high-level visual understanding tasks will thus largely depend on the quality of restoration or enhancement. Yet it remains questionable whether restoration-based approaches would actually boost the visual understanding performance, as the restoration/enhancement step is not optimized towards the target task and may bring in misleading information and artifacts too. For example, a recent line of  researches~\cite{AOD_Net,close_loop,liu2017enhance,liu2018connecting,liu2017image,cheng2017robust,face_reg_track,Vasek-LPR-BMVC2018,Expression_analysis,Recognition_surveillance,prabhu2018u} discuss on the intrinsic interplay relationship of low-level vision and high-level recognition/detection tasks, showing that their goals are not always aligned. 
	
	UG$^{2+}$ Challenge Track 2 aims to evaluate and advance the robustness of object detection algorithms in specific poor-visibility situations, including challenging weather and lighting conditions. We structure Challenge 2 into three sub-challenges.
	Each challenge features a different poor-visibility outdoor condition, and diverse training protocols (paired versus unpaired images, annotated versus unannotated, \textit{etc.}).
	For each sub-challenge, we collect a new benchmark dataset captured in realistic poor-visibility environments with real image artifacts caused by rain, haze, insufficiency of light.
	The specific dataset details and evaluation protocols are illustrated in Section~\ref{sec:ug2_track2_datasets}.
	Comparing with previous works and challenges, our challenge and datasets include the following new features:
	\begin{itemize}
		\item \textbf{Covering Complex Degradation.}
		Our datasets capture images with synthetic and real haze (Challenge 2.1), under-exposure (Challenge 2.2), and rain streaks and raindrops (Challenge 2.3), which provides precious resources to measure the statistics and properties of captured images in these scenes and design effective methods to recover the clean version of these images.
		\item \textbf{Supporting Un/Semi/Full-Supervised Learning.}
		In Challenge 2.1 and 2.2, 
		our proposed datasets include paired and unpaired data to support both full-supervised or semi-supervised (optional for participants) training.
		In Challenge 2.3, there is no training data and testing samples provided. Therefore, it is a zero-shot problem and close to unsupervised learning.
		Therefore, our challenge supports all kinds of learning and provides important materials for future researches.
		\item \textbf{Highly Challenging.}
		In Challenge 2.1 and 2.2, the winners only achieve the results below 65 MAP.
		In Challenge 2.3, no participates achieve the results superior to the baseline results.
		The results show that, our datasets are challenging and there is still large room for further improvement.
		\item \textbf{Full Purpose.} 
		Three sub-challenges support to evaluate the high-level tasks for machine vision.
		The Challenge 2.1 and 2.2 also include the paired data, which can support to evaluate for human vision.
	\end{itemize}
	
	The rest of the article is organized as follows. Section~\ref{sec:related_work} briefly reviews previous works on poor visibility enhancement and visual recognition under adverse conditions as well as the related dataset efforts.
	Section~\ref{sec:ug2_track2_datasets} provides the detailed introduction of our datasets, challenges, evaluation protocols and baseline results.
	Section~\ref{sec:results_analysis} illustrates the competition results and related analysis.
	Section~\ref{sec:insights} summarizes the interesting observations, the reflected insights and briefly discusses the future directions.
	The concluding remarks are provided in Section~\ref{sec:conclusion}.
	
	\section{Related Work}
	\label{sec:related_work}
	\subsection{Datasets}
	Most datasets used for image enhancement/processing mainly targets at evaluating the quantitative (PSNR, SSIM, \textit{etc.}) or qualitative (visual subjective quality) differences of enhanced images \textit{w.r.t.} the ground truths. Some earlier classical datasets include Set5~\cite{Set5}, Set14~\cite{Set14}, and LIVE1~\cite{LIVE}. The numbers of their images are small.
	Subsequent datasets come with more diverse scene content, such as BSD500~\cite{bsds500} and Urban100~\cite{urban100}. The popularity of deep learning methods has increased demand for training and testing data. Therefore, many newer and larger datasets are presented for image and video restoration, such as DIV2K~\cite{DIV2K} and MANGA109~\cite{Manga109} for image super-resolution, PolyU~\cite{PolyU} and Darmstadt~\cite{Darmstadt} for denoising, RawInDark~\cite{SeeInDark} and LOL dataset~\cite{LOL} for low light enhancement, HazeRD~\cite{HazeRD}, OHAZE~\cite{OHAZE} and IHAZE~\cite{IHAZE} for dehazing, Rain100L/H~\cite{yang2017deep} and Rain800~\cite{zhang2017image} for rain streak removal, and RAINDROP~\cite{qian2018attentive} for raindrop removal. However, these datasets provide no integration with subsequent high-level tasks. 
	
	A few works~\cite{SCface,GroupProfiling,VideoSynopsis} make preliminary attempts for event/action understanding, video summarization, or face recognition in unconstrained and potentially degraded environments. The following datasets are collected by aerial vehicles, including VIRAT Video Dataset~\cite{VIRAT} for event recognition, UAV123~\cite{UAV_benchmark} for UAV tracking, and a multi-purpose dataset~\cite{Yao2007IntroductionTA}. In~\cite{UFDD}, an unconstrained Face Detection Dataset (UFDD) is proposed for face detection in adverse condition including weather-based degradations, motion blur, focus blur and several others, containing a total of 6,425 images with 10,897 face annotations. However, few works specifically consider the impacts of image enhancement and object detection/recognition jointly. Prior to this UG$^{2+}$ effort, a number of latest works have taken the first stabs. A large-scale hazy image dataset and a comprehensive study -- REalistic Single Image DEhazing (RESIDE)~\cite{li2019benchmarking} -- including paired synthetic data and unpaired real data is proposed to thoroughly examine visual reconstruction and vision recognition in hazy images. In~\cite{Exdark}, an Exclusively Dark (ExDARK) dataset is proposed with a collection of 7,363 images captured from very low-light environments with 12 object classes annotated on both image class level and local object bounding boxes. In~\cite{li2019single}, the authors present a new large-scale benchmark called RESIDE and a comprehensive study and evaluation of existing single image deraining algorithms, ranging from full-reference metrics, to no-reference metrics, to subjective evaluation and the novel task-driven evaluation. Those datasets and studies shed new light on the comparisons and limitations of state-of-the-art algorithms, and suggest promising future directions. In this work, we follow the footsteps of predecessors to advance the fields by proposing new benchmarks.
	
	\subsection{Poor Visibility Enhancement}
	
	There are numerous algorithms aiming to enhance visibility of the degraded imagery, such as image and video denoising/inpainting \cite{wang2013robust,li2013detection,ren2013context,wang2016d3,liu2018structure}, deblurring \cite{yan2017image,ren2017video,kupyn2018deblurgan,deblurganv2,DAVID}, super-resolution \cite{wang2015learning,wang2015self,liu2017robust,liu2018learning} and interpolation \cite{yu2013multi}. Here we focus on dehazing, low-light condition, and deraining, as in the UG$^{2+}$ Track 2 scope.
	
	\noindent \textbf{Dehazing.} Dehazing methods proposed in an early stage rely on the exploitation of natural image priors and depth statistics, \textit{e.g.} locally constant constraints and decorrelation of the transmission~\cite{single_dehazing}, dark channel prior~\cite{dark_Channel}, color attenuation prior~\cite{color_attenuation_prior}, nonlocal prior~\cite{NL_dehazing}. 
	In \cite{retinex_model,retinex_model2}, Retinex theory is utilized to approximate the spectral properties of object
	surfaces by the ratio of the reflected light.
	Recently, Convolutional Neural Network (CNN)-based methods bring in the new prosperity for dehazing.
	Several methods~\cite{DehazeNet,Ren-ECCV-2016} rely on various CNNs to learn the transmission fully from data. Beyond estimating the haze related variables separately, successive works make their efforts to estimate them in a unified way.
	In~\cite{factorizing,beyesian_defogging}, the authors use a factorial Markov random field that integrates the estimation of transmission and atmosphere light. some researchers focus on the more challenging night-time dehazing problem~\cite{nightdahaze1,nightdahaze2}.
	In addition to image dehazing, AOD-Net~\cite{AOD_Net,li2017all} considers the joint interplay effect of dehazing and object detection in an unified framework.
	The idea is further applied to video dehazing by extending the model into a light-weight video hazing framework~\cite{EVD_Net}. In another recent work~\cite{dehaze_semantic}, the semantic prior is also injected to facilitate video dehazing. 
	
	\noindent \textbf{Low Light Enhancement}. All low-light enhancement methods can be categorized into three ways: hand-crafted methods, Retinex theory-based methods and data-driven methods. Hand-crafted methods explore and apply various image priors to single image low-light enhancement, \textit{e.g.} histogram equalization~\cite{HE1,HE2}.
	Some methods~\cite{dehaze1,dehaze2} regard the inverted low-light images as hazy images, and enhance the visibility by applying dehazing. 
	The retinex theory-based method~\cite{Land77theretinex} is designed to transform the signal components, reflectance and illumination, differently to simultaneously suppress the noise and preserve high-frequency details. Different ways~\cite{Single_scale_retinex,MR} are used to decompose the signal and diverse priors~\cite{Natural_MSR,FU201682,Guo_2017_Lime,Fu_2016_WTV} are applied to realize better light adjustment and noise suppression.
	Li \textit{et al.}~\cite{Li_2017_SRRM} further extended the traditional Retinex model to a robust one with an explicit noise term, and made the first attempt to estimate a noise map out of that model via an alternating direction minimization algorithm. A successive work~\cite{Ren_2018_SD} develops a fast sequential algorithm. Learning based low-light image enhancement methods~\cite{yang_2016_dict,LLNet,MSRNet_2017} have also been studied, where low-light images used for training are synthesized by applying random Gamma transformation on natural normal light images. Some recent works aim to build paired training data from real scenes. In~\cite{SeeInDark}, Chen \textit{et al.} introduced a See-in-the-Dark (SID) dataset of short-exposure low-light raw images with corresponding long-exposure reference raw images. Cai \textit{et al.}~\cite{cai2018learning} built a dataset of under/over-contrast and normal-contrast encoded image pairs, in which the reference normal-contrast images are generated by Multi-Exposure image Fusion or High Dynamic Range algorithms. Recently, Jiang \textit{et al.}~\cite{jiang2019enlightengan} proposed for the first time an unsupervised generative adversarial network, that can be trained without low/normal-light image pairs, yet generalizing nicely and flexibly on various real-world images. 
	
	\noindent \textbf{Deraining}. Single image deraining is a highly ill-posed problem. To address it, many models and priors are used to perform signal separation and texture classification. These models include sparse coding~\cite{ID}, generalized low rank model~\cite{chen2013generalized}, nonlocal mean filter~\cite{nonlocal_derain}, discriminative sparse coding~\cite{luo2015removing}, Gaussian mixture model~\cite{li2016rain}, rain direction prior~\cite{zhang2018density}, transformed low rank model~\cite{Chang_2017_ICCV}. 
	The presence of deep learning has promoted the development of single image deraining. In~\cite{Fu_2017_CVPR,Fu_TIP}, deep networks take the image detail layer as their input.
	Yang \textit{et al.}~\cite{yang2017deep} proposed a deep joint rain detection and removal method to remove heavy rain streaks and accumulation. In~\cite{zhang2018density}, a novel density-aware multi-stream densely connected CNN is proposed for joint rain density estimation and removal. Video deraining can additionally make use of the temporal context and motion information.
	The early works formulate rain streaks with more flexible and intrinsic characteristics, including rain modeling~\cite{garg2006photorealistic,garg2004detection,garg2007vision,garg2005does,zhang2006rain,liu_2009,barnum2010analysis,Santhaseelan_2015,brewer2008using,bossu2011rain,chen2013generalized,Jiang_2017_CVPR}.
	The presence of learning-based method~\cite{Chen_2014,Tripathi_2011,Tripathi_2012,Ren_2017_CVPR,Li_2018_CVPR,Wei_2017_ICCV,Kim_2015_tip}, with improved modeling capacity, brings new progress. The emergence of deep learning-based methods push performance of video deraining to a new level. Chen \textit{et al.}~\cite{Chen_2018_CVPR} integrated superpixel segmentation alignment, and consistency among these segments and CNN-based detail compensation network into a unified framework. Liu \textit{et al.}~\cite{Liu_2018_CVPR} presented a recurrent network integrating rain degradation classification, deraining and background reconstruction.
	
		\begin{table}[t]
		\begin{center}
			\caption{Sub-challenge 2.1: Image and object statistics of the training/validation, and the held-out test sets.}
			\begin{tabular}{c|c|c}
				\hline
				& \#Images & \#Bounding Boxes  \\
				\hline
				\hline
				\textbf{Training/Validation}  & 4,310 & 41,113  \\
				\textbf{Test} (held-out)  & 2,987 & 24,201  \\
				\hline
			\end{tabular}
		\end{center}
		\vspace{-5mm}
	\end{table}
	
	\begin{table}[t]\scriptsize
		\begin{center}
			\caption{Sub-challenge 2.1: Class statistics of the training/validation, and the held-out test sets.}
			\footnotesize
			\begin{tabular}{c|c|c|c|c|c}
				\hline
				Categories & \textit{Car} & \textit{Person} & \textit{Bus} & \textit{Bicycle} & \textit{Motorcycle} \\
				\hline
				\hline
				\textbf{RTTS}  & 25,317 & 11,366 & 2,590 & 698 & 1,232  \\
				\textbf{Test} (held-out)  & 18,074 & 1,562 & 536 & 225 & 3,804 \\
				\hline
			\end{tabular}
		\end{center}
		\vspace{-5mm}
	\end{table}
	
	\subsection{Visual Recognition under Adverse Conditions
	}
	
	A real-world visual detection/recognition system needs to handle a complex mixture of both low-quality and high-quality images.
	It is commonly observed that, mild degradations, \textit{e.g.} small noises, scaling with small factors, lead to almost no change of recognition performance. However, once the degradation level passes a certain threshold, there will be an unneglected or even very significant effect on system performance.
	In~\cite{32pixel}, Torralba \textit{et al.} showed that, there will be a significant performance drop in object and scene recognition when the image resolution is reduced to 32$\times$32 pixels.
	In~\cite{16pixel}, the boundary where the face recognition performance is largely degraded is 16$\times$16 pixels.
	Karahan \textit{et al.}~\cite{std} found Gaussian noise with its standard deviation ranging from 10 to 20 will cause a rapid performance decline.	In~\cite{face_impact}, more impacts of contrast, brightness, sharpness, and out-of-focus on face recognition are analyzed.
	
	In the era of deep learning, some methods~\cite{remove_shaking,scsr,arcnn} attempt to first enhance the input image and then forward the output into a classifier. 
	However, this separate consideration of enhancement may not benefit the successive recognition task, because the first stage may incur artifacts which will damage the second stage recognition.
	In~\cite{16pixel,joint_first}, class-specific features are extracted as a prior to be incorporated into the restoration model.
	In~\cite{close_loop}, Zhang \textit{et al.} developed a joint image restoration and recognition method based on sparse representation prior, which constrains the identity of the test image and guides better reconstruction and recognition. In~\cite{AOD_Net}, Li \textit{et al.} considered dehazing and object detection jointly. 
	These two stage joint optimization methods achieve better performance than previous one-stage methods. In~\cite{wang2016studying,liu2017enhance}, the joint optimization pipeline for low-resolution recognition is examined. In~\cite{liu2017image,liu2018connecting}, Liu \textit{et al.} discussed the impact of denoising for semantic segmentation and advocated their mutual optimization. Lately, in~\cite{vidalmata2019bridging}, the algorithmic impact of enhancement algorithms for both visual quality and automatic object recognition is thoroughly examined, on a real image set with highly compound degradations. In our work, we take a further step to consider the joint enhancement and detection in bad weather environments. Three large-scale datasets are collected to inspire new ideas and novel methods in the related fields. 
	
	\begin{figure*}[htbp]
		\centering
		\subfigure{
			\includegraphics[width=4.5cm]{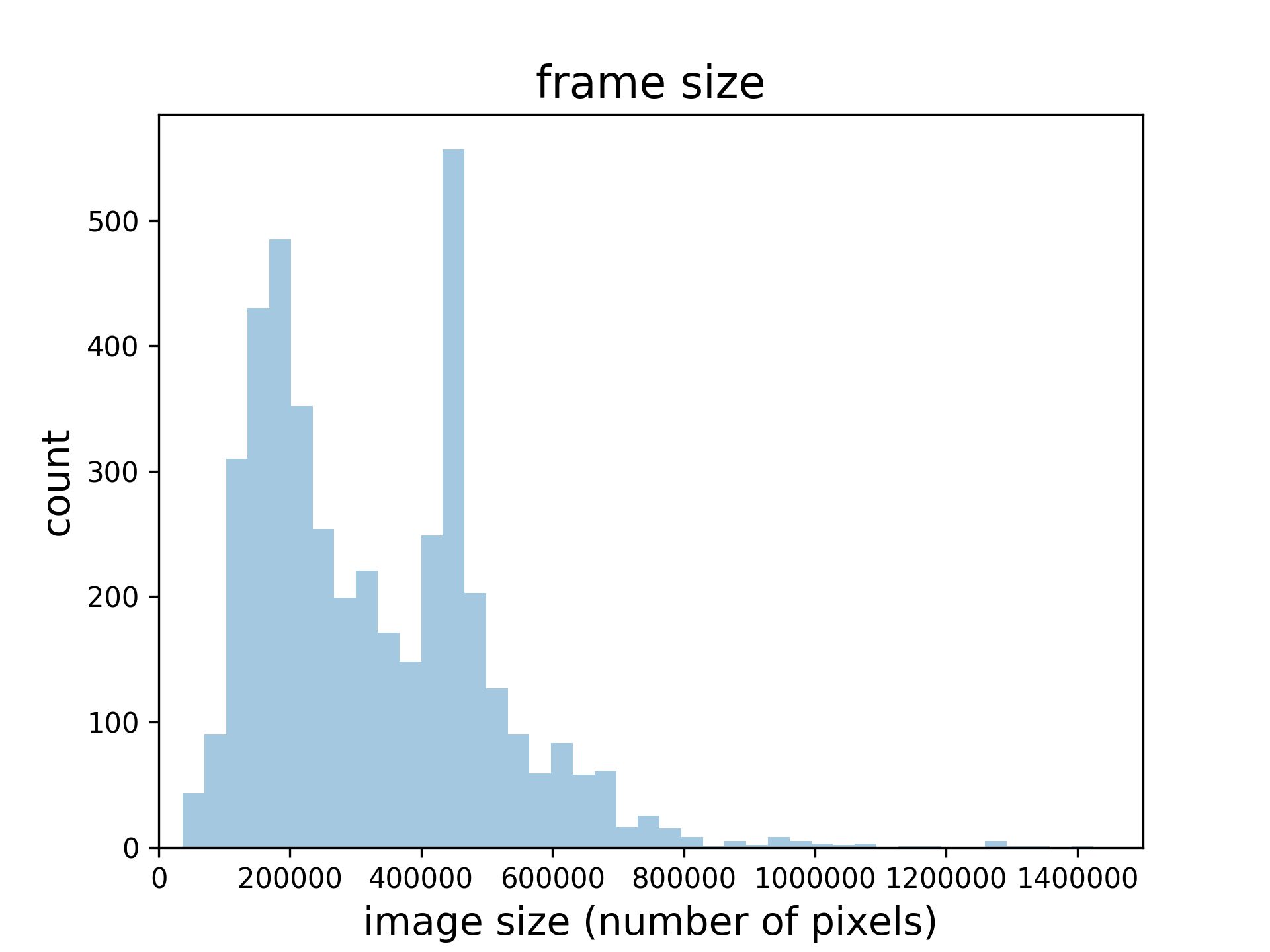}
		} \hspace{-6mm}
		\subfigure{
			\includegraphics[width=4.5cm]{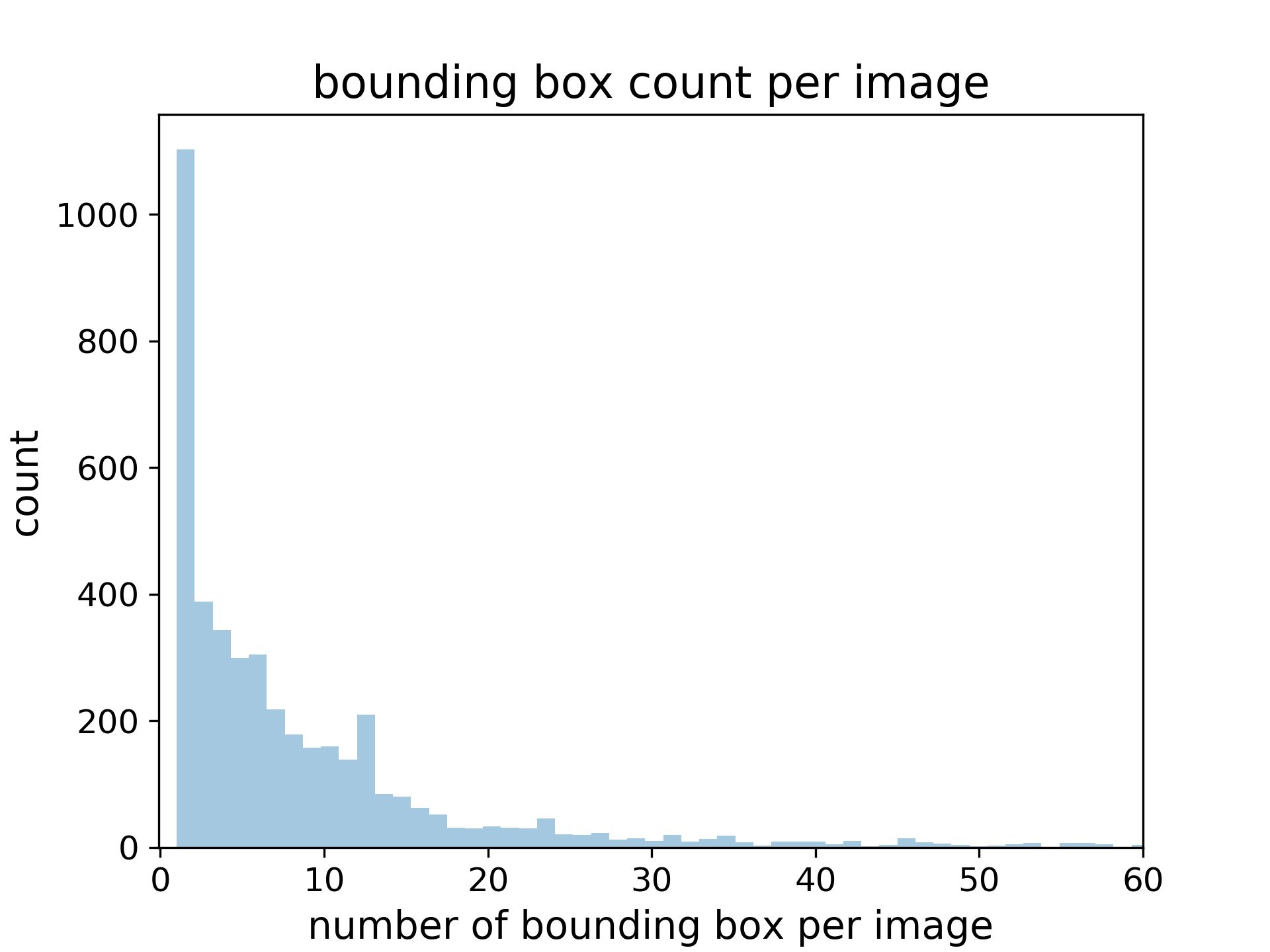}
		} \hspace{-6mm}
		\subfigure{
			\includegraphics[width=4.5cm]{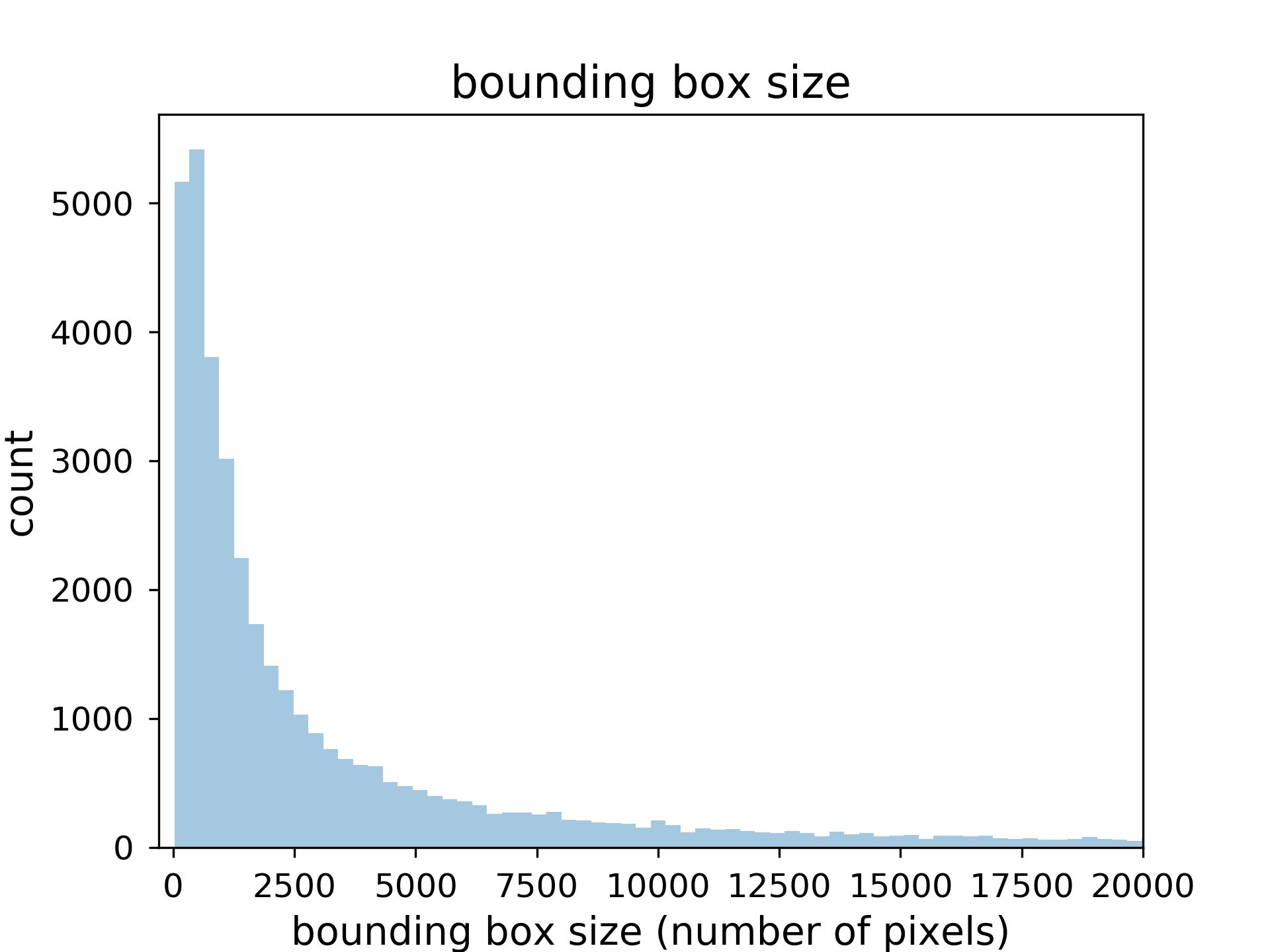}
		} \hspace{-6mm}
		\subfigure{
			\includegraphics[width=4.5cm]{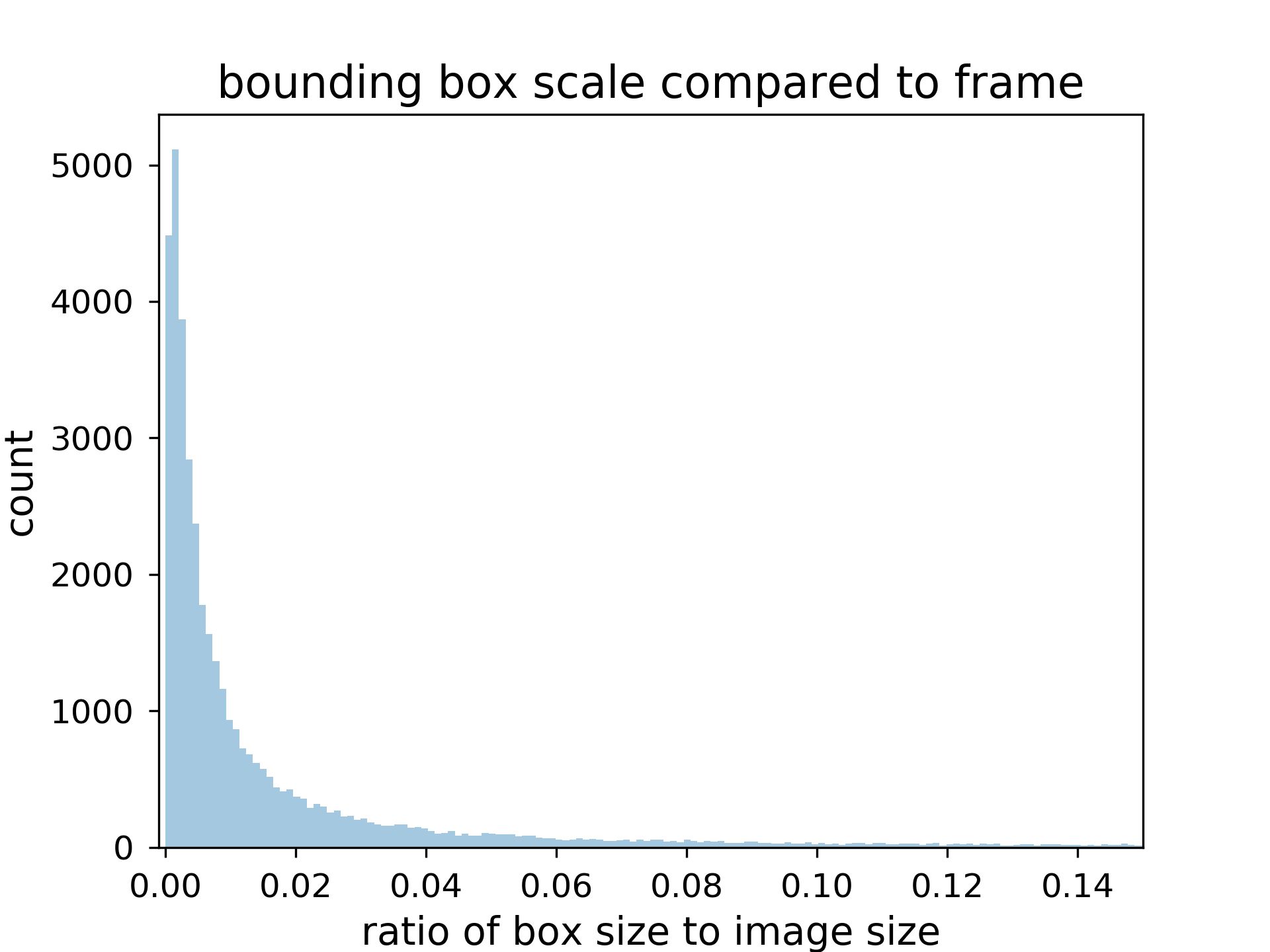}
		} \hspace{-6mm}
		
		\subfigure{
			\includegraphics[width=4.5cm]{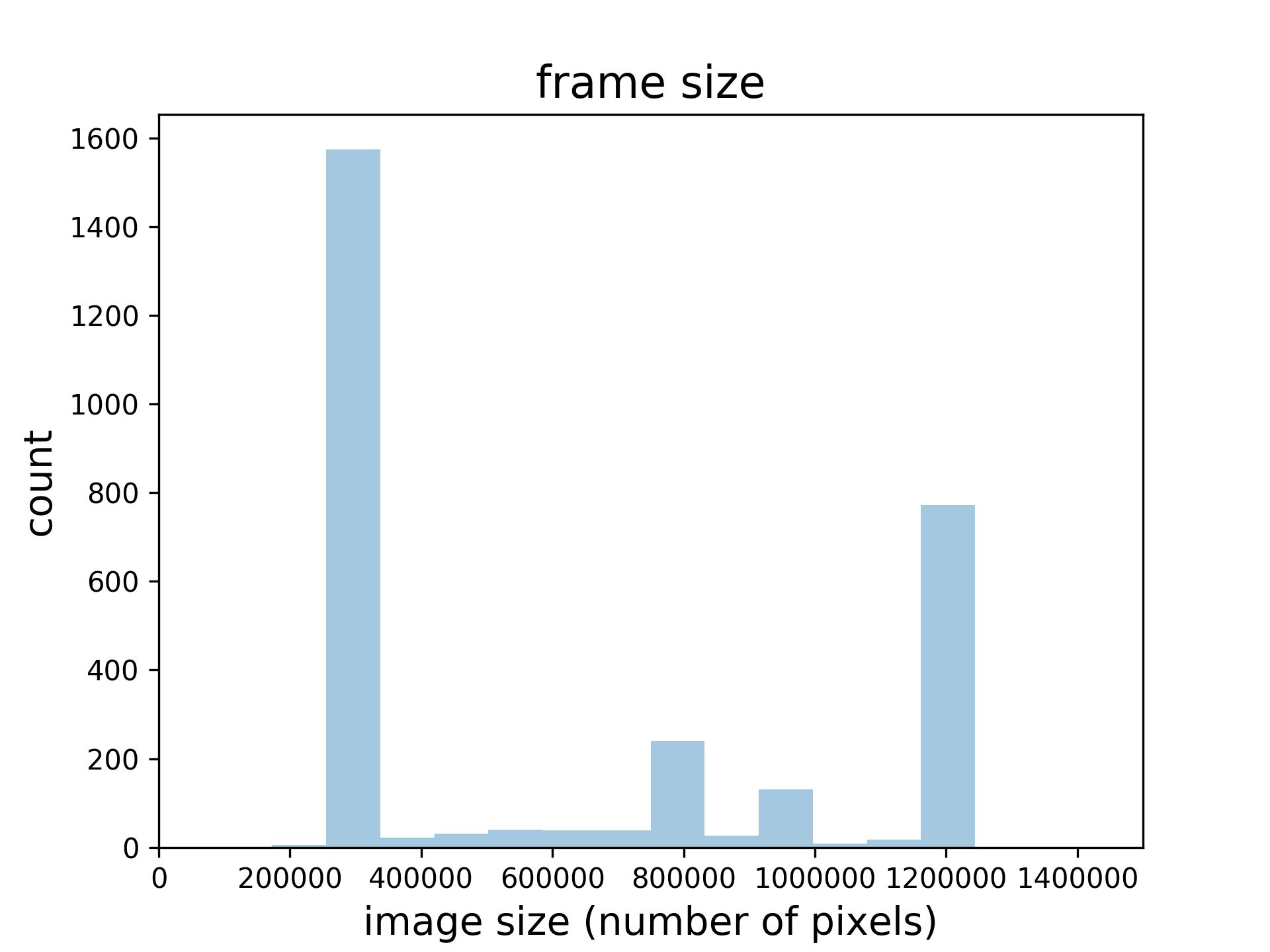}
		} \hspace{-6mm}
		\subfigure{
			\includegraphics[width=4.5cm]{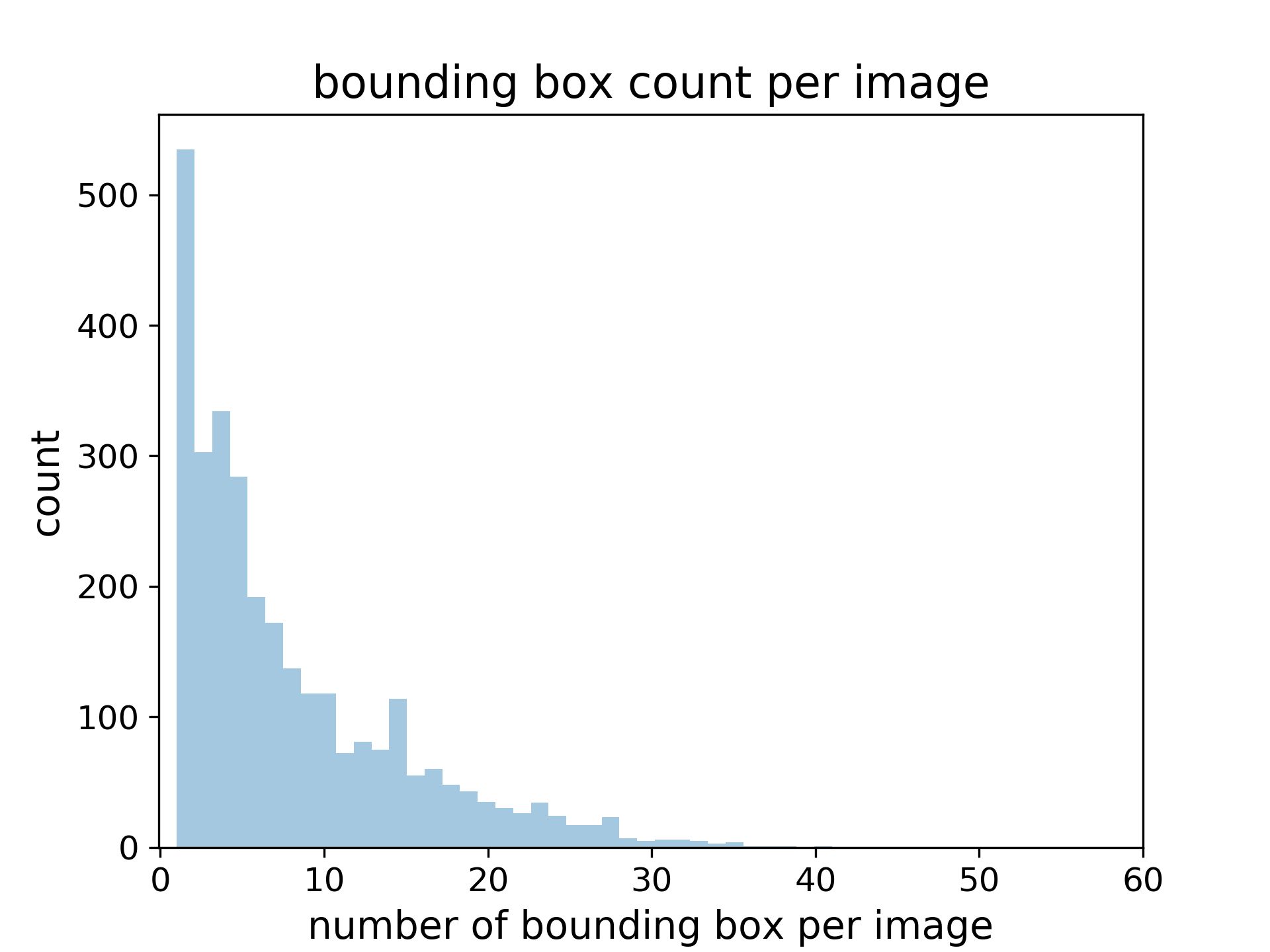}
		} \hspace{-6mm}
		\subfigure{
			\includegraphics[width=4.5cm]{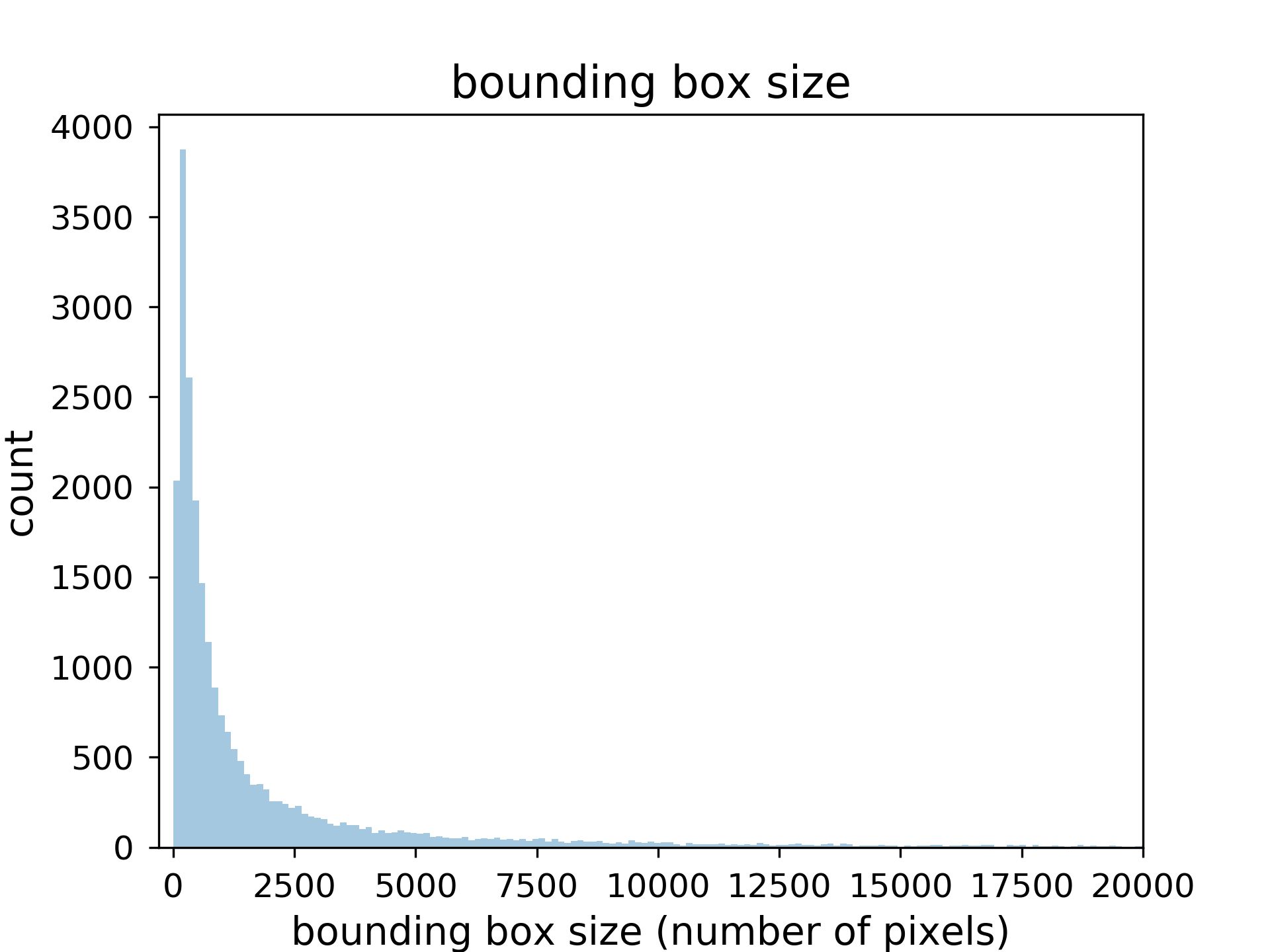}
		} \hspace{-6mm}
		\subfigure{
			\includegraphics[width=4.5cm]{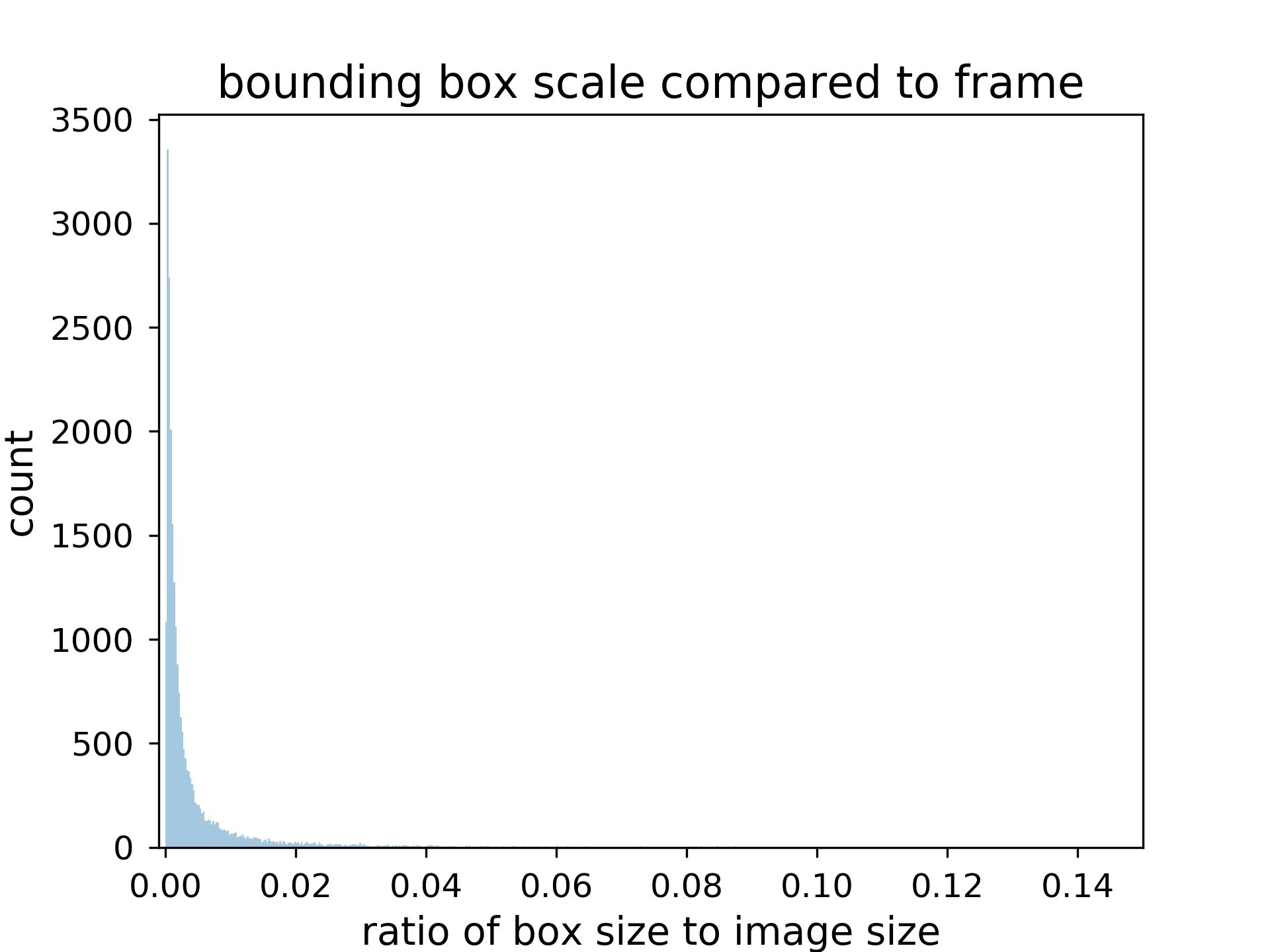}
		} \hspace{-6mm}
		\caption{Sub-challenge 2.1: Basic statistics on the training/validation set (the top row) and the held out test set (the bottom row). The first column shows the image size distribution (number of pixels per image),  The second column the bounding box count distribution (number of bounding boxes per image), the third column the bounding box size distribution (number of pixels per bounding box), and the last column the ratios of bounding box size compared to frame size.}
		\label{fig:hazy_stat}
	\end{figure*}

	\section{Introduction of UG$^{2+}$ Track 2 Datasets}
	\label{sec:ug2_track2_datasets}
	
	\subsection{(Semi-)Supervised Object Detection in the Haze}
	
	\begin{figure}[t!]
		\centering
		\centering
		\subfigure
		{	
			\includegraphics[width=4cm]{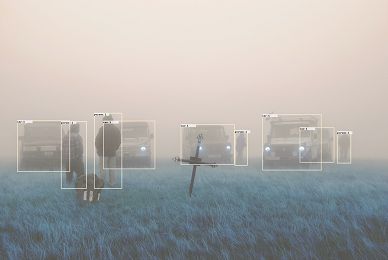}
		} \hspace{-2mm}
		\subfigure
		{	
			\includegraphics[width=4cm]{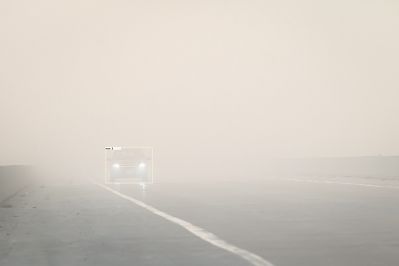}
		}\\ \vspace{-2mm}
		\subfigure
		{	
			\includegraphics[width=4cm]{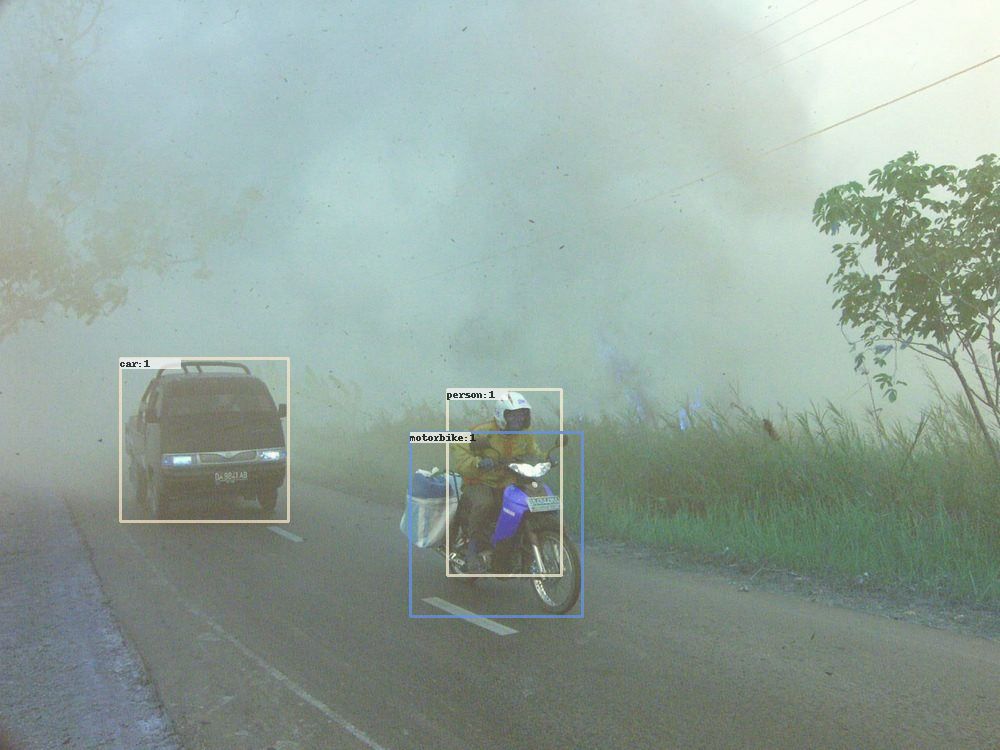}
		} \hspace{-2mm}
		\subfigure
		{	
			\includegraphics[width=4cm]{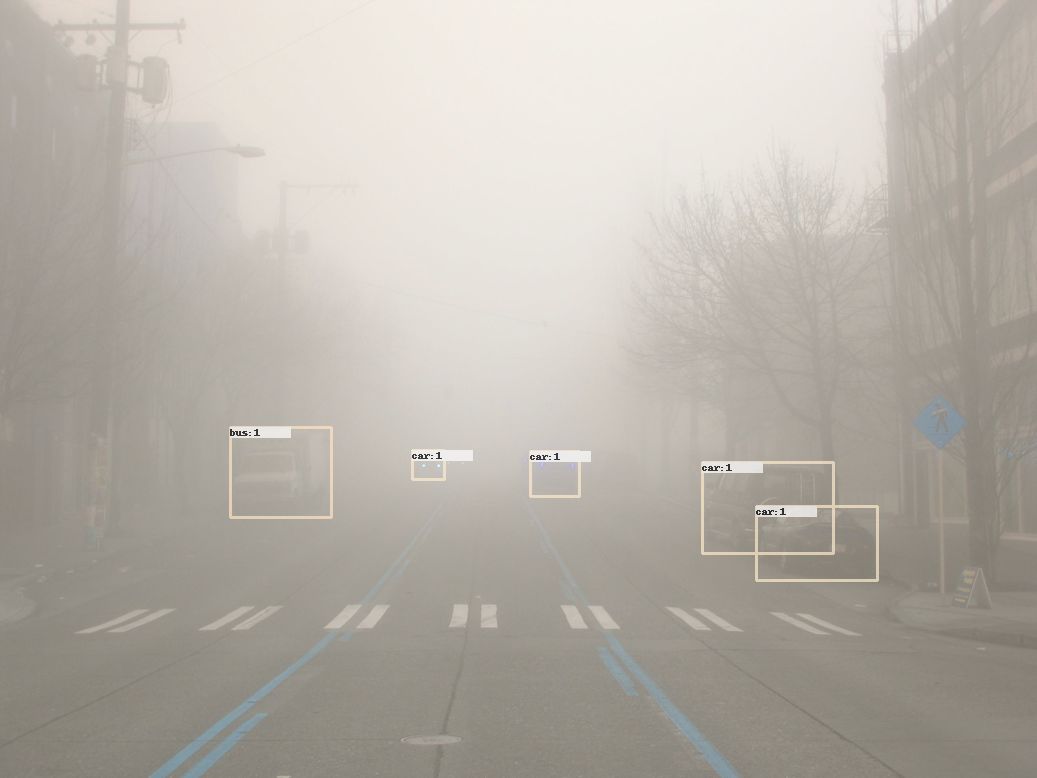}
		}\\
		\caption{Sub-challenge 2.1: Examples of images in training/validation set (\textit{i.e.}, RESIDE  RTTS \cite{li2019benchmarking}).}
		\label{fig:haze_rtts_example}
	\end{figure}
	
	\begin{figure}[t!]
		\centering
		\subfigure
		{	
			\includegraphics[width=4cm]{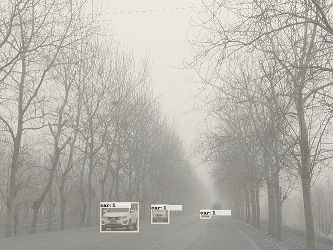}
		} \hspace{-2mm}
		\subfigure
		{	
			\includegraphics[width=4cm]{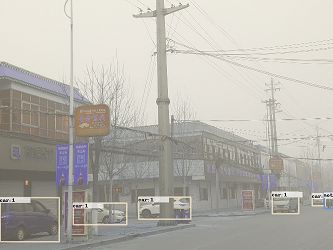}
		}\\ \vspace{-2mm}
		\subfigure
		{	
			\includegraphics[width=4.3cm]{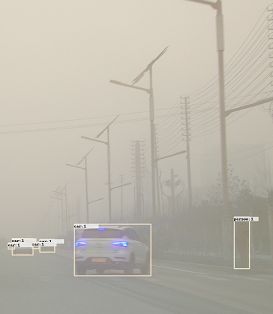}
		} \hspace{-2mm}
		\subfigure
		{	
			\includegraphics[width=3.7cm]{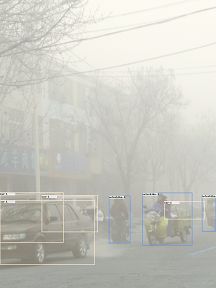}
		}\\ 
		\caption{Sub-challenge 2.1: Examples of images in the held-out test set.}
		\label{fig:haze_test_example}
	\end{figure}
	
	In Sub-challenge 2.1, we use the 4,322 annotated real-world hazy images  of the RESIDE RTTS set \cite{li2019benchmarking} as the training and/or validation sets (the split is up to the participants). Five categories of objects (car, bus, bicycle, motorcycle, pedestrian) are labeled with tight bounding boxes. We provide another 4,807 unannotated real-world hazy images collected from the same traffic camera sources, for the possible usage of semi-supervised training. The participants can optionally use pre-trained models (\textit{e.g.}, on ImageNet or COCO), or external data. But if any pre-trained model, self-synthesized or self-collected data is used, that must be explicitly mentioned in their submissions, and the participants must ensure all their used data to be public available at the time of challenge submission, for reproduciblity purposes. 
	
	There is a held-out test set of 2,987 real-world hazy images, collected from the same sources, with the same classes of objected annotated. Fig.~\ref{fig:hazy_stat} shows the basic statistics of the RTTS set and the held-out set. 
	The held-out test set has a similar distribution of number of bounding boxes per image, bounding box size and relative scale of bounding boxes to input images compared to the RTTS set, but has relatively larger image size. Samples from RTTS set and held-out set can be found in Fig.~\ref{fig:haze_rtts_example} and Fig.~\ref{fig:haze_test_example}.
	
	\subsection{(Semi-)Supervised Face Detection in the Low Light Condition}
	
	In Sub-challenge 2.2, we use our self-curated DARK FACE dataset. It is composed of 10,000 images (6,000 for training and validation, and 4,000 for testing) taken in under-exposure condition where human faces are annotated by human with bounding boxes; and 9,000 images taken with the same equipment in the similar environment without human annotations. Additionally, we provide a unique set of 789 paired low-light/normal-light images captured in controllable real lighting conditions (but unnecessarily containing faces), which can be optionally used as parts of the training data. The training and evaluation set includes 43,849 annotated faces and the held-out test set includes 37,711
	annotated faces. Table~\ref{tab:compare} presents a summary of the dataset.
	
	\begin{figure}[t]
		\centering
		\subfigure
		{	
			\includegraphics[width=4cm]{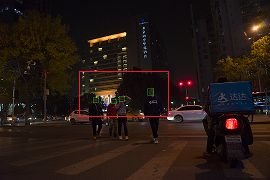}
		} \hspace{-2mm}
		\subfigure
		{	
			\includegraphics[width=4cm]{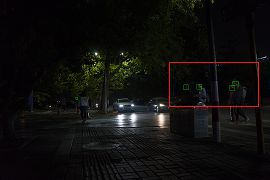}
		}\\ \vspace{-2mm}
		\subfigure
		{	
			\includegraphics[width=4cm]{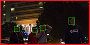}
		} \hspace{-2mm}
		\subfigure
		{	
			\includegraphics[width=4cm]{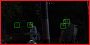}
		}\\ \vspace{-1mm}	
		\subfigure
		{	
			\includegraphics[width=4cm]{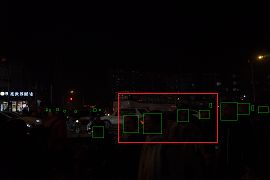}
		} \hspace{-2mm}
		\subfigure
		{	
			\includegraphics[width=4cm]{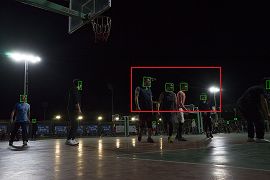}
		}
		\\ \vspace{-2mm}
		\subfigure
		{	
			\includegraphics[width=4cm]{./figure_example/diversity/dark_face_diversity_1_part1.jpg}
		} \hspace{-2mm}
		\subfigure
		{	
			\includegraphics[width=4cm]{./figure_example/diversity/dark_face_diversity_2_part1.jpg}
		}	
		\caption{
			Sub-challenge 2.2: DARK FACE has a high degree of variability in scale, pose, occlusion, appearance and illumination. \wh{The face regions in the red boxes are zoomed-in for better viewing.}
		}
		\label{fig:diversity}
	\end{figure}

	\begin{figure*}[t]
		\centering
		\subfigure[FN in Train]
		{	
			\includegraphics[width=4cm]{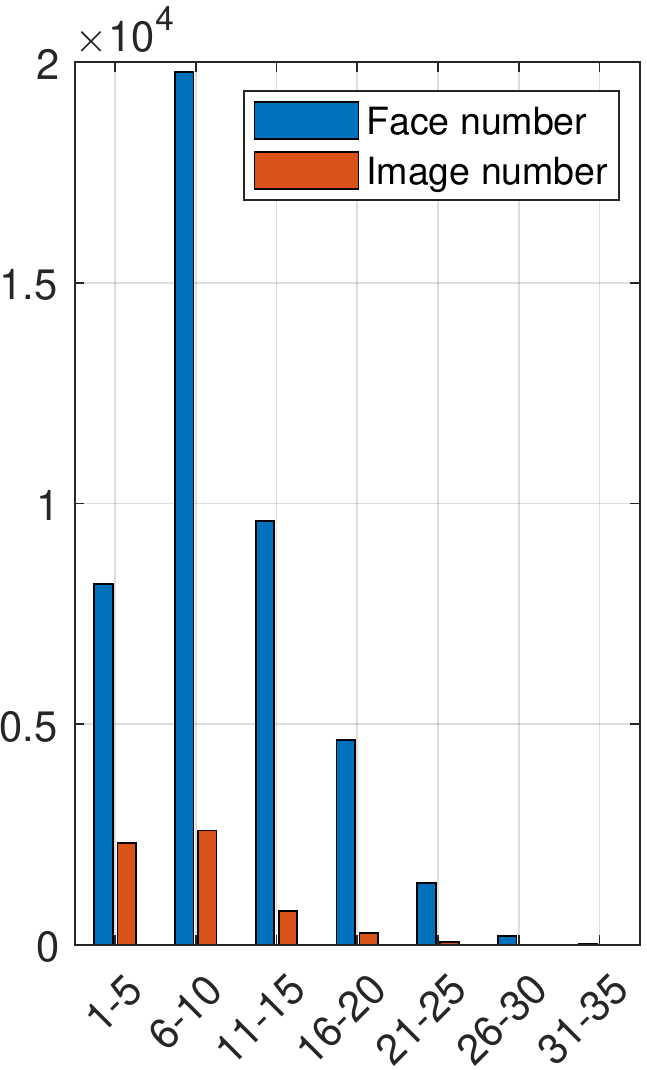}
		}
		\hspace{1mm}
		\subfigure[FN in Test]
		{	
			\includegraphics[width=4cm]{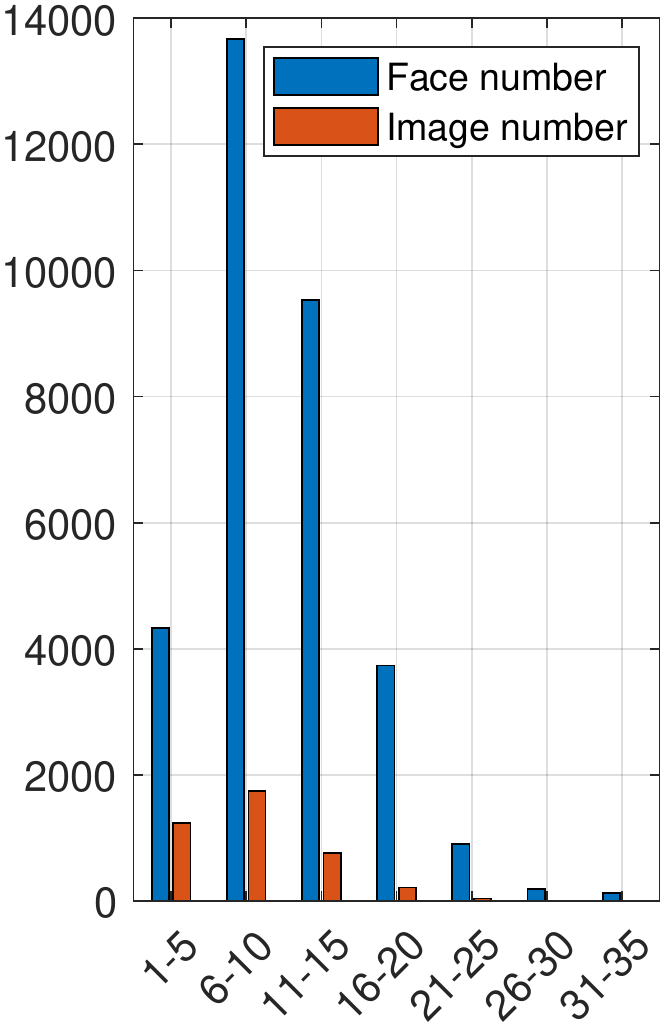}
		}
		\hspace{1mm}	
		\subfigure[FR in Train]
		{	
			\includegraphics[width=4cm]{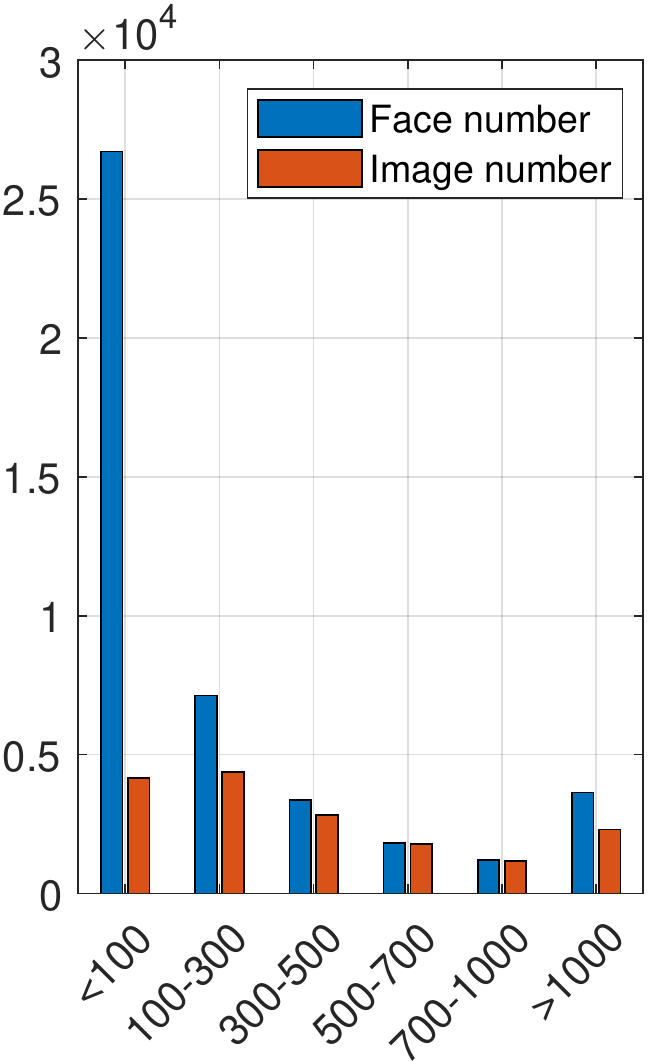}
		}
		\hspace{1mm}
		\subfigure[FR in Test]
		{	
			\includegraphics[width=4cm]{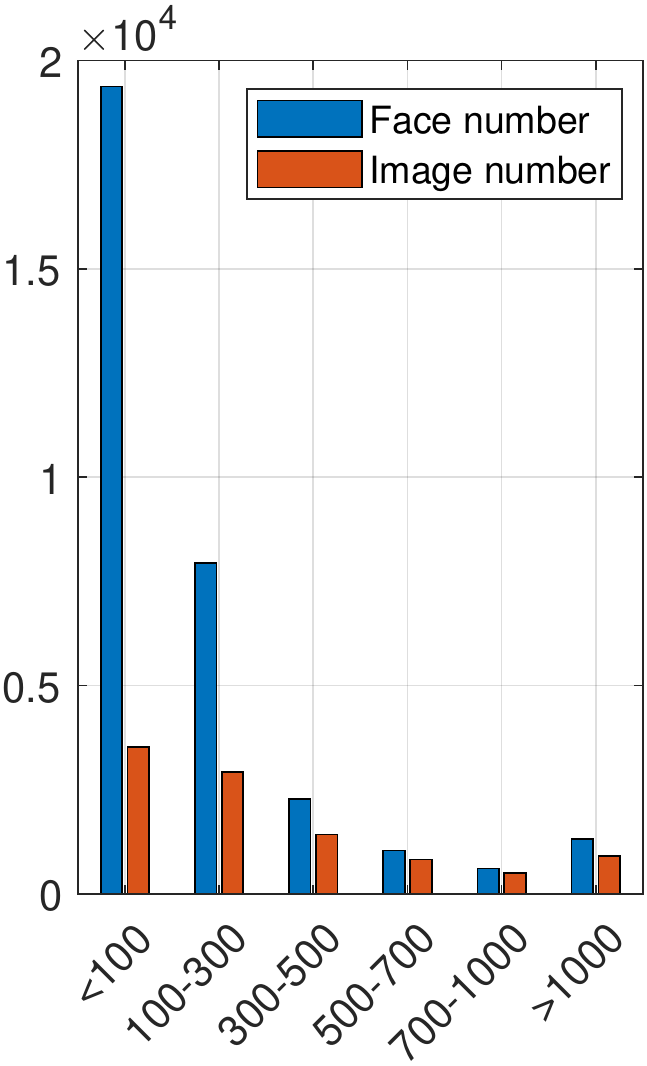}
		}
		\caption{Sub-challenge 2.2: Face resolution (FR) and face number (FN) distribution in DARK FACE collections. Image number denotes the number of images belonging to a certain category. Face number denotes the summation number of faces belonging to a certain category.}
		\label{fig:face_distribution}
	\end{figure*}

	\begin{table}[t]
		\centering
		\caption{Sub-challenge 2.2: Comparison of low-light image understanding datasets.}
		\label{tab:compare}	
		\begin{tabular}{c|cc|cc}
			\hline
			\multirow{2}{*}{\textbf{Dataset}} & \multicolumn{2}{c|}{Training} & \multicolumn{2}{c}{Testing} \\ \cline{2-5} 
			& \#Image        & \#Face       & \#Image       & \#Face       \\ \hline\hline
			ExDark                            & 400            & -            & 209           & -            \\ 
			UFDD                              & -              & -            & 612           & -            \\ 
			\textbf{DarkFace}                 & 6,000           & 43,849        & 4,000          &  37,711     \\ \hline
		\end{tabular}
	\end{table}

	\noindent \textbf{Collection and annotation}. This collection consists of images recorded from Digital Single Lens Reflexes, specifically Sony $\alpha6000$ and $\alpha7$ E-mount cameras with different capturing parameters on several busy streets around Beijing, where faces of various scales and poses are captured. 
	The images in this collection are open source content tagged with an Attribution-NonCommercial-NoDerivatives 4.0 International license\footnote{https://www.jet.org.za/clearinghouse/projects/primted/resources/creative-commons-licence}. 
	The resolution of these images is 1080 $\times$720 (down-sampled from 6K $\times$ 4K). After filtering out those without sufficient information (lacking faces, too dark to see anything, \textit{etc.}), we select 10,000 images for human annotation. 
	The bounding boxes are labeled for all the recognizable faces in our collection. We make the bounding boxes tightly around the forehead, chin, and cheek, using the LabelImg Toolbox\footnote{https://github.com/tzutalin/labelImg}. If a face is occluded, we only label the exposed skin region. If most of a face is occluded, we ignore it. For this collection, we observe commonly seen degradations in addition to under-exposure, such as intensive noise.
	The face number and resolution range distribution are displayed in Fig~\ref{fig:face_distribution}. Each annotated image contains 1-34 human faces. The face resolutions in these images range from 1$\times$2 to 335$\times$296. The resolution of most faces in our dataset is below 300 pixel$^2$ and the the face number mostly falls into the range $[1,20]$. 	

	\begin{figure}[htbp]\footnotesize
	\centering
	\tabcolsep 1pt
	\begin{tabular}{@{}cc@{}}
		\includegraphics[width=.99\linewidth]{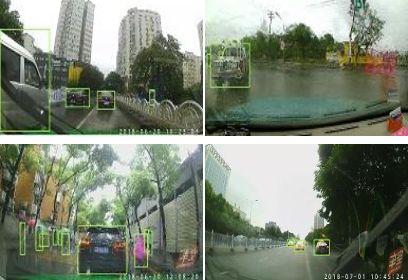}	
	\end{tabular}
	\caption{Sub-challenge 2.3: Example images from the held-out test set.}
	\vspace{-2mm}
	\label{fig:DRR}
\end{figure} 

	\begin{table}\footnotesize
	\begin{center}
		\caption{Sub-challenge 2.3: Object statistics in the held-out test set.}
		\label{RID-RIS}
		\begin{tabular}{c|c|c|c|c|c}
			\hline
			Categories             &  \textit{Car}  & \textit{Person}  & \textit{Bus} & \textit{Bicycle} & \textit{Motorcycle}  \\
			\hline
			\hline
			\textbf{Test Set}  & 7332  & 1135 & 613 & 268 & 968 \\
			\hline 
		\end{tabular}                   
	\end{center}
	\vspace{-5mm}
\end{table}

		\begin{figure*}[htbp]
	\centering
	\subfigure
	{	
		\includegraphics[width=3.5cm]{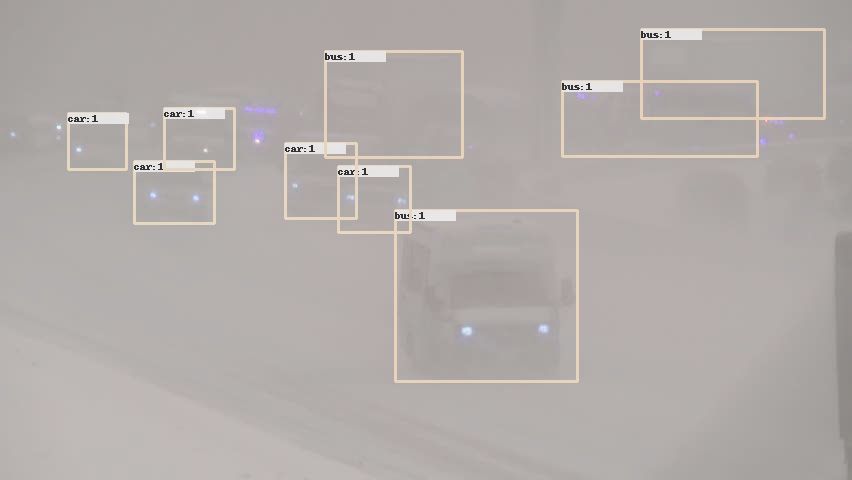}
		\includegraphics[width=3.5cm]{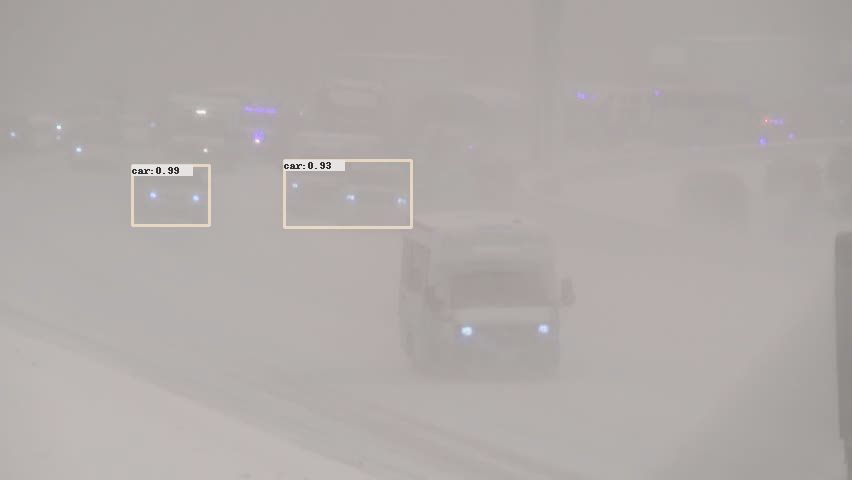}
		\includegraphics[width=3.5cm]{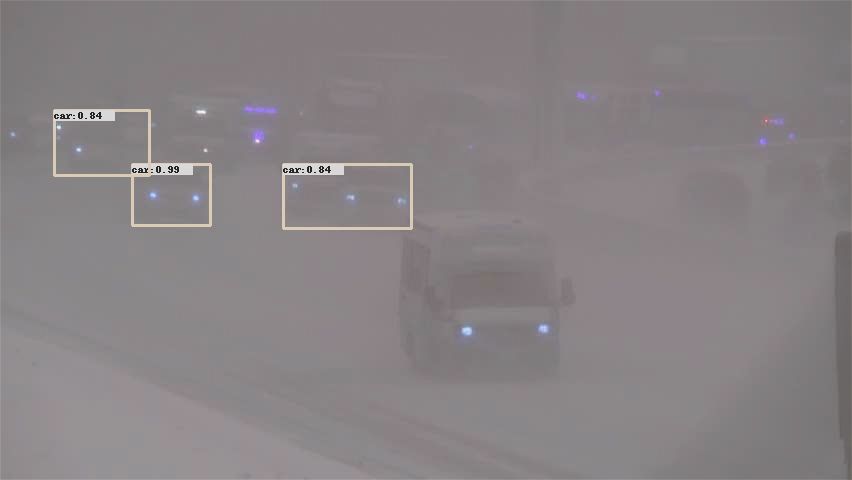}
		\includegraphics[width=3.5cm]{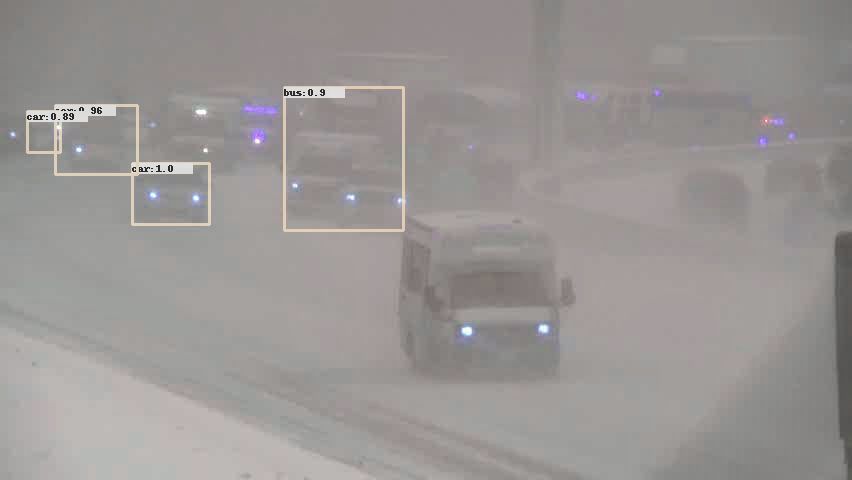}
		\includegraphics[width=3.5cm]{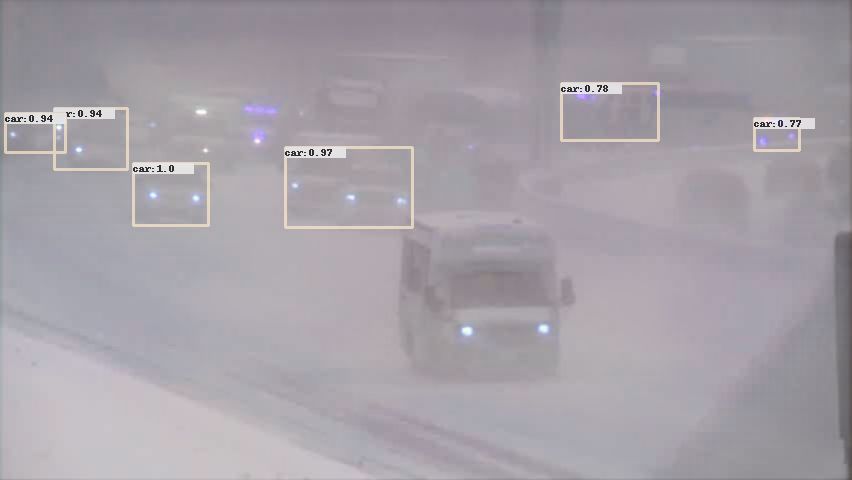}
	} \vspace{-2mm}
	\subfigure
	{	
		\includegraphics[width=3.5cm]{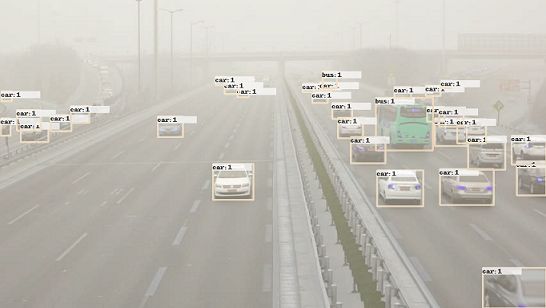}
		\includegraphics[width=3.5cm]{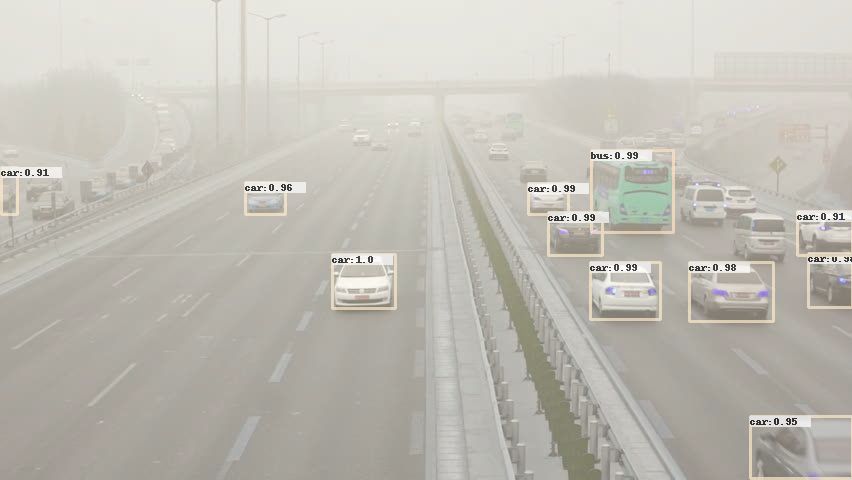}
		\includegraphics[width=3.5cm]{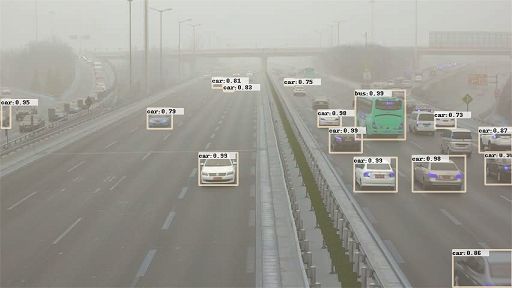}
		\includegraphics[width=3.5cm]{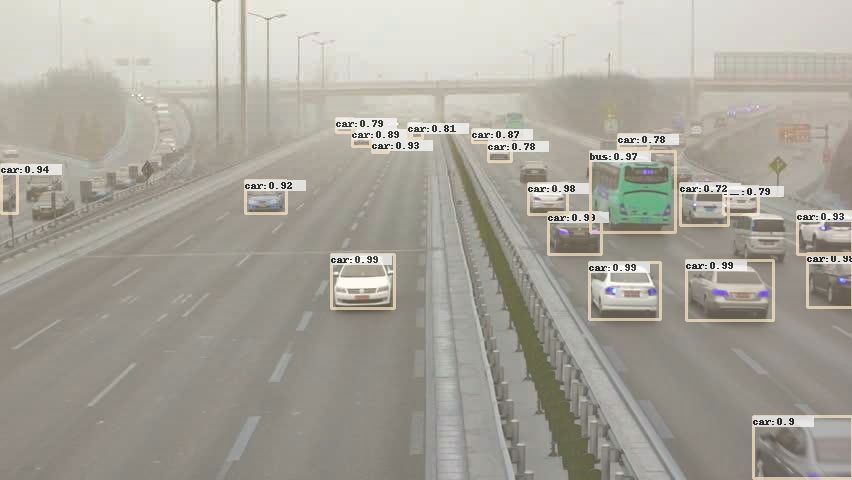}
		\includegraphics[width=3.5cm]{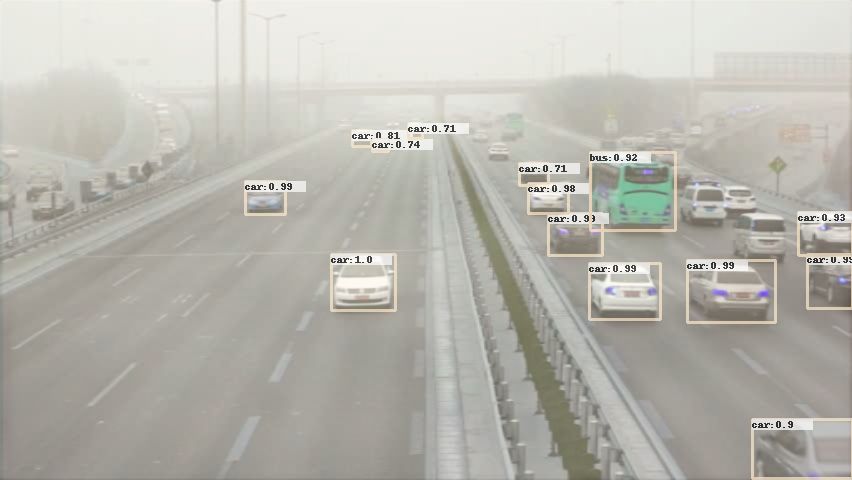}
	} \vspace{-2mm}
	\\
	\caption{Examples of object detection of hazy images and dehazed images on RESIDE  RTTS  set. The first column displays the ground truth bounding boxes on hazy images, the second column displays detected bounding box on hazy image using pretrained Mask R-CNN, the right three columns display Mask R-CNN detected bounding boxed on dehazed images using AOD-Net, MSCNN, DCPDN correspondingly.} 
	\label{fig:haze_maskrcnn_dehaze}
\end{figure*}

	\begin{figure}[t]
	\centering
	\subfigure
	{	
		\includegraphics[width=8.5cm]{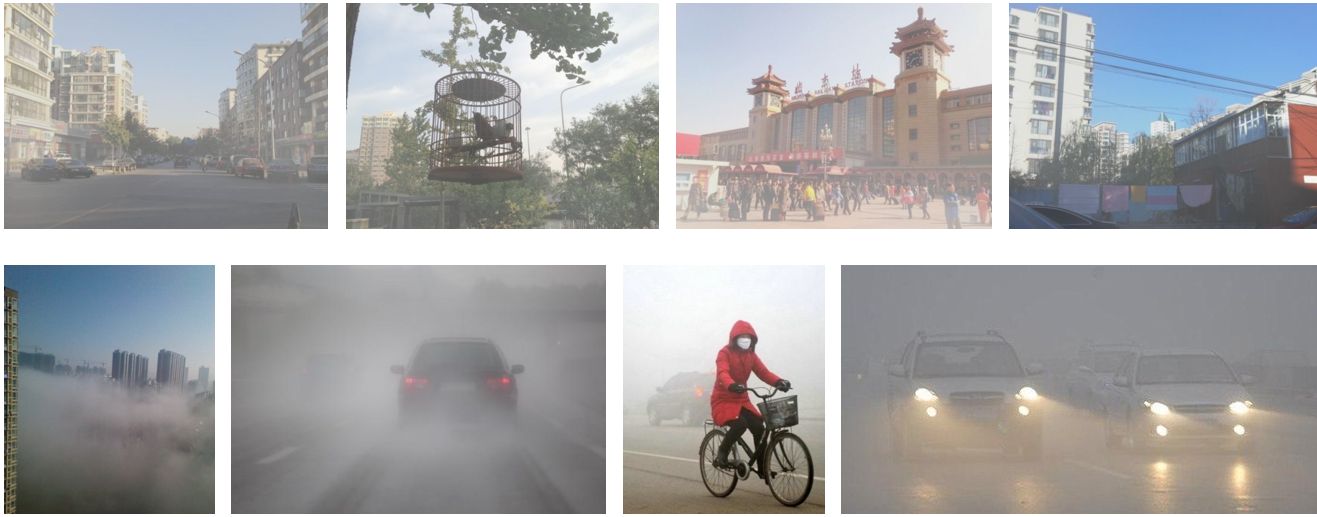}
	}
	\caption{Compared with synthetic hazy images in OTS dataset (at the top panel), the haze layers of real hazy images (at the bottom panel) in RTTS dataset might be uncorrelated to the scene depth and have more variations for objects with various depths in single hazy images.} 
	\label{fig:haze_visual}
\end{figure}
	
	\subsection{Zero-Shot Object Detection with Raindrop Occlusions}

	In Sub-challenge 2.3, we release 1,010 pairs of realistic raindrop images and corresponding clean ground-truths, collected from the real scenes as described in \cite{qian2018attentive}, as the training and/or validation sets. Our held-out test set contains 2,495 real rainy images from high-resolution driving videos. As shown in Fig.~\ref{fig:DRR}, all images are contaminated by raindrops on camera lens. They are captured in diverse real traffic locations and scenes during multiple drives. We label bounding boxes for selected traffic objects: car, person, bus, bicycle, and motorcycle, which commonly appear on the roads of all images. Most images are of 1920 $\times$ 990 resolution, with a few exceptions of 4023 $\times$ 3024 resolution. The participants are free to use pre-trained models (\textit{e.g.}, ImageNet or COCO) or external data. But if any pre-trained model, self-synthesized or self-collected data is used, that must be explicitly mentioned in their submissions, and the participants must ensure their used data to be public available at the time of challenge submission, for reproducibility purposes.
	
	\subsection{Ranking Criterion}
	
	The ranking criteria will be the Mean average precision (mAP) on each held-out test set, with a default Interception-of-Union (IoU) threshold as 0.5. If the ratio of the intersection of a detected region with an annotated object is greater than 0.5, a score of 1 is assigned to the detected region, and 0 otherwise. When mAPs with IoU as 0.5 are equal, the mAPs with higher IoUs (0.6, 0.7, 0.8) will be compared sequentially.
	
	\section{Baseline Results and Analysis}
	
	For all three sub-challenges, we report results by \textbf{cascading off-the-shelf enhancement methods and popular pre-trained detectors}. 
	There has been no joint training performed, hence the baseline numbers are in no way very competitive. We expect to see much performance boosts over the baselines from the competition participants.

	\subsection{Sub-challenge 2.1 Baseline Results}
	
	\subsubsection{Baseline Composition}
	We test four state-of-the-art object detectors: 
	(1) \textbf{Mask R-CNN}\footnote{https://github.com/matterport/Mask\_RCNN} \cite{he2017mask}; 
	(2) \textbf{RetinaNet}\footnote{https://github.com/fizyr/keras-retinanet} \cite{lin2018focal}; and 
	(3) \textbf{YOLO-V3}\footnote{https://github.com/ayooshkathuria/pytorch-yolo-v3} \cite{redmon2018yolov3};
	(4) Feature Pyramid Network\footnote{https://github.com/DetectionTeamUCAS/FPN\_Tensorflow} \textbf{(FPN)} \cite{lin2017feature}. We also try three state-of-the-art dehazing approaches: 
	(a) \textbf{AOD-Net}\footnote{https://github.com/Boyiliee/AOD-Net} \cite{AOD_Net}; 
	(b) Multi-Scale Convolutional Neural Network (\textbf{MSCNN})\footnote{https://github.com/rwenqi/Multi-scale-CNN-Dehazing} \cite{Ren-ECCV-2016}; 
	(c) Densely Connected Pyramid Dehazing Network (\textbf{DCPDN})\footnote{https://github.com/hezhangsprinter/DCPDN} \cite{zhang2018densely}. 
	All dehazing models adopt officially released versions.
	
	\begin{table*}[!ht]
		\begin{center}
			\caption{Detection results (mAP) on the RTTS (train/validation dataset) and held-out test sets.}
			\label{dehaze_res}
			\footnotesize
			\begin{tabular}{c|c|c|c|c|c|c}
				\hline
				\multicolumn{3}{c|}{mAP} & hazy & AOD-Net \cite{AOD_Net} & DCPDN \cite{Ren-ECCV-2016} & MSCNN \cite{zhang2018densely} \\
				\hline
				\multirow{18}{*}{\textbf{validation}}
				& \multirow{6}{*}{$\ast$ RetinaNet \cite{lin2018focal}}
				& Person        & 55.85 & 54.93 & 56.70 & \textbf{58.07} \\
				&& Car          & 41.19 & 37.61 & 42.68 & \textbf{42.77} \\
				&& Bicycle      & 39.61 & 37.80 & \textbf{36.96} & 38.16 \\
				&& Motorcycle   & 27.37 & 23.31 & \textbf{29.18} & 29.01 \\
				&& Bus          & 16.88 & 15.70 & 16.34 & \textbf{18.34} \\
				&& mAP          & 36.18 & 33.87 & 36.37 & \textbf{37.27} \\
				\cline{2-7}
				& \multirow{6}{*}{$\ast$ Mask R-CNN \cite{he2017mask}}
				& Person        & 67.52 & 66.71 & 67.18 & \textbf{69.23} \\
				&& Car          & 48.93 & 47.76 & \textbf{52.37} & 51.93 \\
				&& Bicycle      & \textbf{40.81} & 39.66 & 40.40 & 40.42 \\
				&& Motorcycle   & 33.78 & 26.71 & \textbf{34.58} & 31.38 \\
				&& Bus          & 18.11 & 16.91 & 18.25 & \textbf{18.42} \\
				&& mAP          & 41.83 & 39.55 & \textbf{42.56} & 42.28 \\
				\cline{2-7}
				& \multirow{6}{*}{$\ast$ YOLO-V3 \cite{redmon2018yolov3}}
				& Person        & 60.81 & 60.21 & 60.42 & \textbf{61.56} \\
				&& Car          & 47.84 & 47.32 & 48.17 & \textbf{49.75} \\
				&& Bicycle      & 41.03 & \textbf{42.22} & 40.18 & 42.01 \\
				&& Motorcycle   & 39.29 & 37.55 & 38.17 & \textbf{41.11} \\
				&& Bus          & 23.71 & 20.91 & \textbf{23.35} & 23.15 \\
				&& mAP          & 42.54 & 41.64 & 42.06 & \textbf{43.52} \\
				\cline{2-7}
				& \multirow{6}{*}{$\diamond$ FPN \cite{lin2017feature}} 
				& Person        & 51.85 & 52.35 & 51.04 & \textbf{54.50} \\
				&& Car          & 37.48 & 36.05 & 37.19 & \textbf{38.88} \\
				&& Bicycle      & 35.31 & 35.93 & 32.57 & \textbf{37.01} \\
				&& Motorcycle   & 23.65 & 21.07 & 22.97 & \textbf{23.86} \\
				&& Bus          & 12.95 & 13.68 & 12.07 & \textbf{15.83} \\
				&& mAP          & 32.25 & 31.82 & 31.17 & \textbf{34.02} \\
				\hline
				\multirow{18}{*}{\textbf{test}}
				& \multirow{6}{*}{RetinaNet}
				& Person        & 17.64 & 18.23 & 16.65 & \textbf{19.34} \\
				&& Car          & 31.41 & 29.30 & 33.31 & \textbf{32.97} \\
				&& Bicycle      & 0.42  & \textbf{0.84}  & 0.38  & 0.75 \\
				&& Motorcycle   & 1.69  & 1.37  & 1.93  & \textbf{2.03} \\
				&& Bus          & 12.77 & 13.70 & 12.07 & \textbf{15.82} \\
				&& mAP          & 12.79 & 12.69 & 12.87 & \textbf{14.18} \\
				\cline{2-7}
				& \multirow{6}{*}{Mask R-CNN}
				& Person        & 25.60 & 26.63 & 24.59 & \textbf{27.94} \\
				&& Car          & 39.31 & 39.71 & \textbf{42.76} & 42.57 \\
				&& Bicycle      & \textbf{0.64}  & 0.52  & 0.22  & 0.37 \\
				&& Motorcycle   & \textbf{3.37}  & 2.81  & 2.83  & 2.99 \\
				&& Bus          & 15.66 & 15.41 & \textbf{16.69} & 16.55 \\
				&& mAP          & 16.92 & 17.02 & 17.42 & \textbf{18.09} \\
				\cline{2-7}
				& \multirow{6}{*}{YOLO-V3}
				& Person        & 20.64 & 21.41 & 21.42 & \textbf{22.11} \\
				&& Car          & 34.68 & 33.90 & 34.52 & \textbf{35.93} \\
				&& Bicycle      & 0.50  & 0.38  & \textbf{0.98}  & 0.57 \\
				&& Motorcycle   & 4.26  & 4.10  & 4.72  & \textbf{5.27} \\
				&& Bus          & 13.55 & 14.35 & 13.75 & \textbf{15.04} \\
				&& mAP          & 14.69 & 14.83 & 15.08 & \textbf{15.78} \\
				\cline{2-7}
				& \multirow{6}{*}{FPN} 
				& Person        & 12.65 & 12.57 & 11.13 & \textbf{14.19} \\
				&& Car          & 30.54 & 31.24 & 27.81 & \textbf{32.68} \\
				&& Bicycle      & \textbf{1.91}  & 0.39  & 1.12  & 0.97 \\
				&& Motorcycle   & \textbf{2.25}  & 1.7   & 1.96  & 1.89 \\
				&& Bus          & 6.08  & 7.93  & 7.39  & \textbf{8.31} \\
				&& mAP          & 10.69 & 10.77 & 9.88  & \textbf{11.61} \\
				\hline
			\end{tabular}
			\begin{tablenotes}
				\centering
				\footnotesize
				\item[$\ast$] $\ast$ RetinaNet, Mask R-CNN and YOLO-V3 are pretrained on Microsoft COCO dataset.
				\item[$\diamond$] $\diamond$ FPN using ResNet-101 backbone is pretrained on the PASCAL Visual Object Classes (VOC) dataset.
			\end{tablenotes}
		\end{center}
		\vspace{-8mm}
	\end{table*}	
	
	\begin{table}[t]
		\scriptsize
		\caption{Comparison of human and machine vision quality different methods achieve.}
		\label{tab:dehazing}
		\begin{tabular}{cccccc}
			\hline
			Metric               & Dataset                     & Baseline   & MSCNN  & AOD-Net & DCPDN  \\
			\hline
			SSIM                 & TestA {[}140{]}             & -          & 0.8203 & 0.8842  & 0.956  \\
			SSIM                 & TestB {[}140{]}             & -          & 0.7724 & 0.8325  & 0.8746 \\
			\hline
			\multirow{4}{*}{MAP} & \multirow{4}{*}{UG2.1-Test} & RetinaNet  & 14.18  & 12.69   & 12.87  \\
			\cline{3-6}
			&                             & Mask R-CNN & 18.09  & 17.02   & 17.42  \\
			\cline{3-6}
			&                             & YOLO-V3    & 15.78  & 14.83   & 15.08  \\
			\cline{3-6}
			&                             & FPN        & 11.61  & 10.77   & 9.88   \\
			\hline
		\end{tabular}
	\end{table}
	
	\subsubsection{Results and Analysis}
	\wh{We evaluate the} object detection performance on the original hazy images of RESIDE  RTTS  set using Mask R-CNN. The detectrons are pretrained on Microsoft COCO, a large-scale object detection, segmentation, and captioning dataset. \wh{The} detailed detection performance on the five objects can be found in Table \ref{dehaze_res}. Results show that without preprocessing or dehazing, the object detectors pretrained on clean images fail to predict a large amount of objects in the hazy image. The overall detection performance has a mAP of only 41.83\% using Mask R-CNN and 42.54\% using YOLO-V3. Among all the five object categories, person has the highest detection AP, while bus has the lowest AP. 
	
	We also compare the validation and test set performance in Table. \ref{dehaze_res}. One possible reason for the performance gap between validation and test sets is that the bounding box size of the latter is smaller compared to the former, as showed in Fig.~\ref{fig:hazy_stat} as well as visualized in Fig.~\ref{fig:haze_maskrcnn_dehaze}.
	
	Besides, we analyze the difference between the synthetic haze/rain images and those in real applications. The haze image is generated from the model:
	\begin{align}
	I=Jt+A(1-t),
	\end{align}
	where $I$ is the observed hazy image, $J$ is the scene reliance to be recovered. $A$ denotes the global atmospheric light, and $t$ is the transmission matrix assumed correlated with the scene depth. For synthesis hazy images as shown at the top panel of Fig.~\ref{fig:haze_visual}, $t$ is strictly inferred from the results of depth estimation. Therefore, the haze is distributed more homogeneously among different regions and objects, as existing depth estimation techniques are not adaptive enough to accurately estimate the fine-grained depth of each object. Comparatively, the haze layers of real hazy images as shown in the bottom panel of Fig.~\ref{fig:haze_visual} might be uncorrelated to the scene depth or have more variations for objects with various depths in single hazy images. Besides, the scattering and refraction of light under real hazy condition is different to that in clear environment. The glow effects in front of the vehicles in haze (\textit{e.g.} last examples) makes the vehicle recognition more difficult using pretrained object detection algorithms.
	
	\subsubsection{Effect of Dehazing}
	We further evaluate the current state-of-the-art dehazing approaches on hazy dataset, with pre-trained detectors subsequently applied without tuning or adaptation. Fig.~\ref{fig:haze_maskrcnn_dehaze} shows two examples that dehazing algorithms can improve not only the visual quality of the images but also the detection accuracies. More detection results are included in Table. \ref{dehaze_res}. Detection mAPs of dehazed images using DCPDN and MSCNN approaches are 1\% higher on average compared to those of hazy images. 
	Eventually, the choice of pre-trained detectors seem to also matter here: Mask R-CNN outperforms the other two detectors on both validation and test sets, before and after dehazing. 
	
	Furthermore, as reported in~\cite{zhang2018densely} and Table~\ref{tab:dehazing}, DCPDN has the best SSIM scores while MSCNN has the worst visual quality. However, the detection performance of MSCNN is much better than that of DCPDN.

	\subsection{Sub-challenge 2.2 Baseline Results}
	
	\subsubsection{Baseline Composition} 
	We test four state-of-the-art deep face detectors: 
	(1) Dual Shot Face Detector (\textbf{DSFD})~\cite{li2018dsfd}\footnote{https://github.com/TencentYoutuResearch/FaceDetection-DSFD};
	(2) \textbf{Pyramidbox}~\cite{Tang_2018_ECCV}\footnote{https://github.com/EricZgw/PyramidBox};
	(3) Single Stage Headless Face Detector (\textbf{SSH})~\cite{ssh}\footnote{https://github.com/mahyarnajibi/SSH.git};
	(4) \textbf{Faster RCNN}~\cite{Jiang2017FaceDW}\footnote{https://github.com/playerkk/face-py-faster-rcnn}.
	
	\vspace{1mm}
	We also include seven state-of-the-art algorithms for light/contrast enhancement: 
	(a) Bio-Inspired Multi-Exposure Fusion (\textbf{BIMEF})~\cite{BIMEF}\footnote{https://github.com/baidut/BIMEF};
	(b) \textbf{Dehazing}~\cite{dong2011fast}$^{14}$;
	(c) Low-light IMage Enhancement (\textbf{LIME})~\cite{Guo_2017_Lime}\footnote{https://sites.google.com/view/xjguo/lime};
	(d) \textbf{MF}~\cite{FU201682}$^{14}$;
	(e) Multi-Scale Retinex (\textbf{MSR)}~\cite{MR}$^{14}$;
	(f) Joint Enhancement and Denoising (\textbf{JED})~\cite{Ren_2018_SD}\footnote{https://github.com/tonghelen/JED-Method};
	(g) \textbf{RetinexNet}~\cite{LOL}\footnote{https://github.com/weichen582/RetinexNet}.
	
	\subsubsection{Results and Analysis}
\begin{figure}[t]
		\centering
		\subfigure{
			\includegraphics[width=4cm]{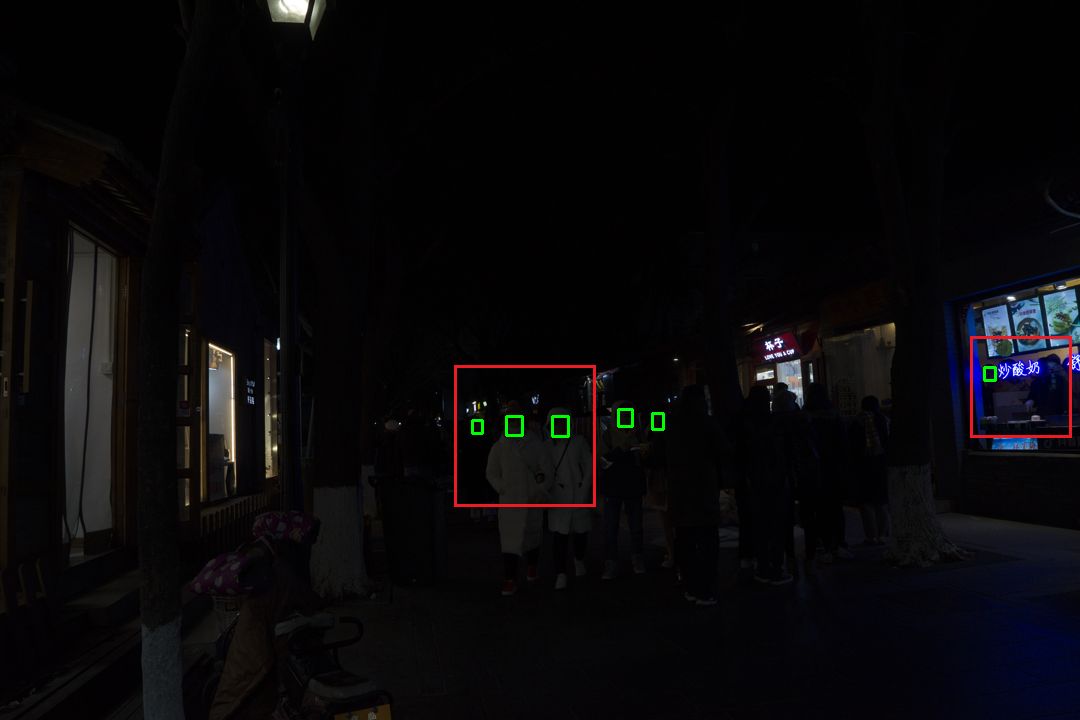}
		}\hspace{-1mm}
		\subfigure{
			\includegraphics[width=4cm]{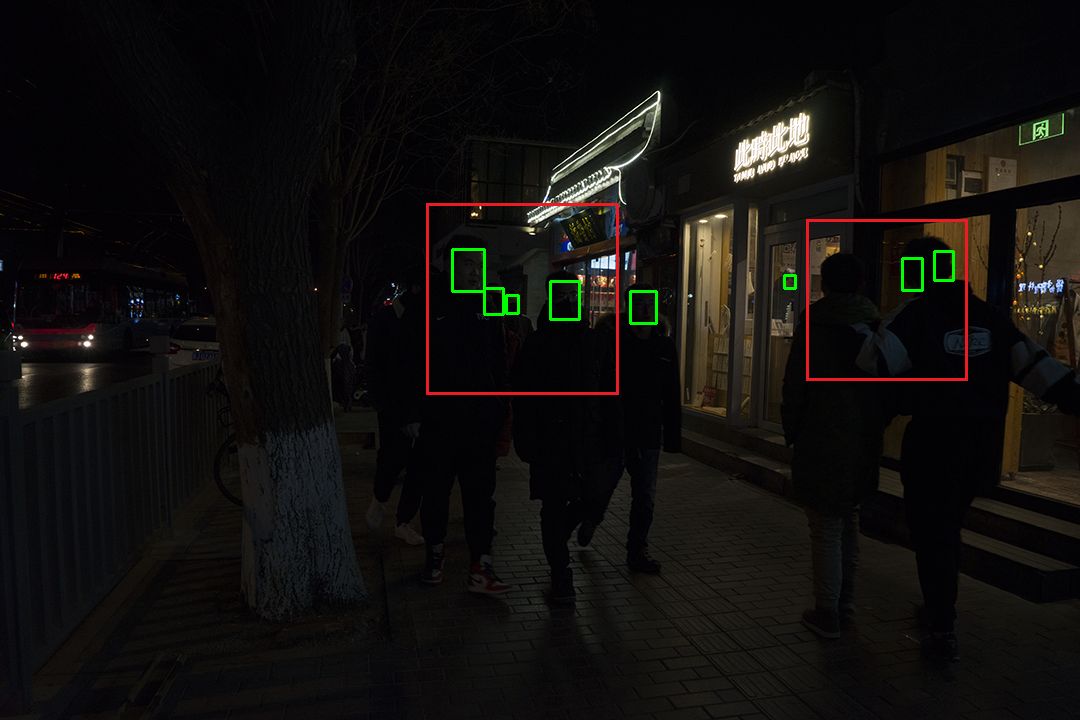}
		}\\ \vspace{-2mm}
		\subfigure{
			\includegraphics[width=1.9cm]{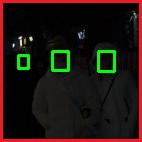}
		}\hspace{-1.5mm}
		\subfigure{
			\includegraphics[width=1.9cm]{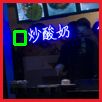}
		}\hspace{-1mm}
		\subfigure{
			\includegraphics[width=1.9cm]{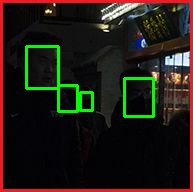}
		}\hspace{-1.5mm}
		\subfigure{
			\includegraphics[width=1.9cm]{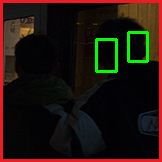}
		}
		\vspace{-2mm}	
		\caption{Sample face detection results of pretrained baseline on the original images of the proposed DARK FACE dataset.
		\wh{The face regions in the red boxes are zoomed-in for better viewing.}}
		\vspace{-3mm}
		\label{fig:Face_pre_results_sub}
	\end{figure}

	Fig.~\ref{fig:Face_after_results}~(a) depicts the precision-recall curves of the original face detection methods, without enhancement. The baseline methods are trained on WIDER FACE~\cite{yang2016wider}\footnote{http://shuoyang1213.me/WIDERFACE/}, a large dataset with large scale variations in diversified factors and conditions. The results demonstrate that without proper pre-processing or adaptation, the state-of-the-art methods cannot achieve desirable detection rates on DARK FACE. Result examples are illustrated in Fig.~\ref{fig:Face_pre_results_sub}. The evidences may imply that previous face datasets, though covering variations in poses, appearances, scale, \textit{et al.}, are still insufficient to capture the facial features in the highly under-exposed condition. 
	
	\begin{figure}[htbp]
		\centering
		\subfigure
		{	
			\includegraphics[width=8.5cm]{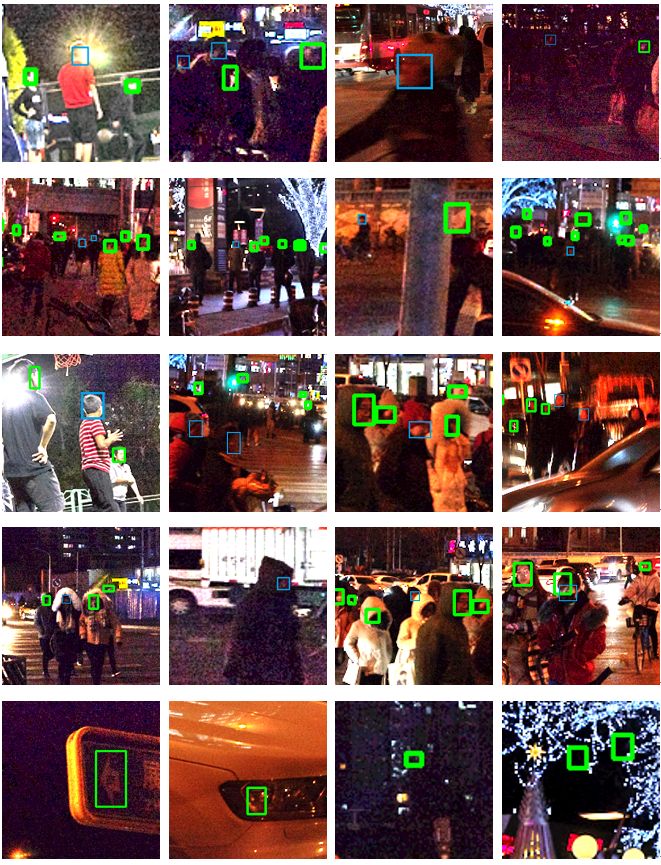}
		}
		\caption{Failure case analysis even with the model being trained with the proposed training set. Green Box: detection results by the baseline DSFD. Blue boxes: ground truths that are not detected.} 
		\label{fig:dark_face_failure}
	\end{figure}
	
	We have showed some failure cases of Sub-challenge 2.2. In these results, the faces are falsely detected (false positives and false negatives) due to heavy degradation, small scale, pose variation, occlusion, \textit{etc}.
	As show in Fig.~\ref{fig:dark_face_failure}, due to the above mentioned reasons, there are a certain amount of false negative samples caused by heavy degradation, small scale, pose variation, occlusion at the top four panels, respectively, and false positive samples at the bottom panel. DSFD is used as the baseline method. For better visibility, the results shown here are processed by LIME.
	
	\subsubsection{Effect of Enhancement}
\begin{figure*}[htbp!]
		\centering
		\subfigure[Original]{
			\includegraphics[width=5.6cm]{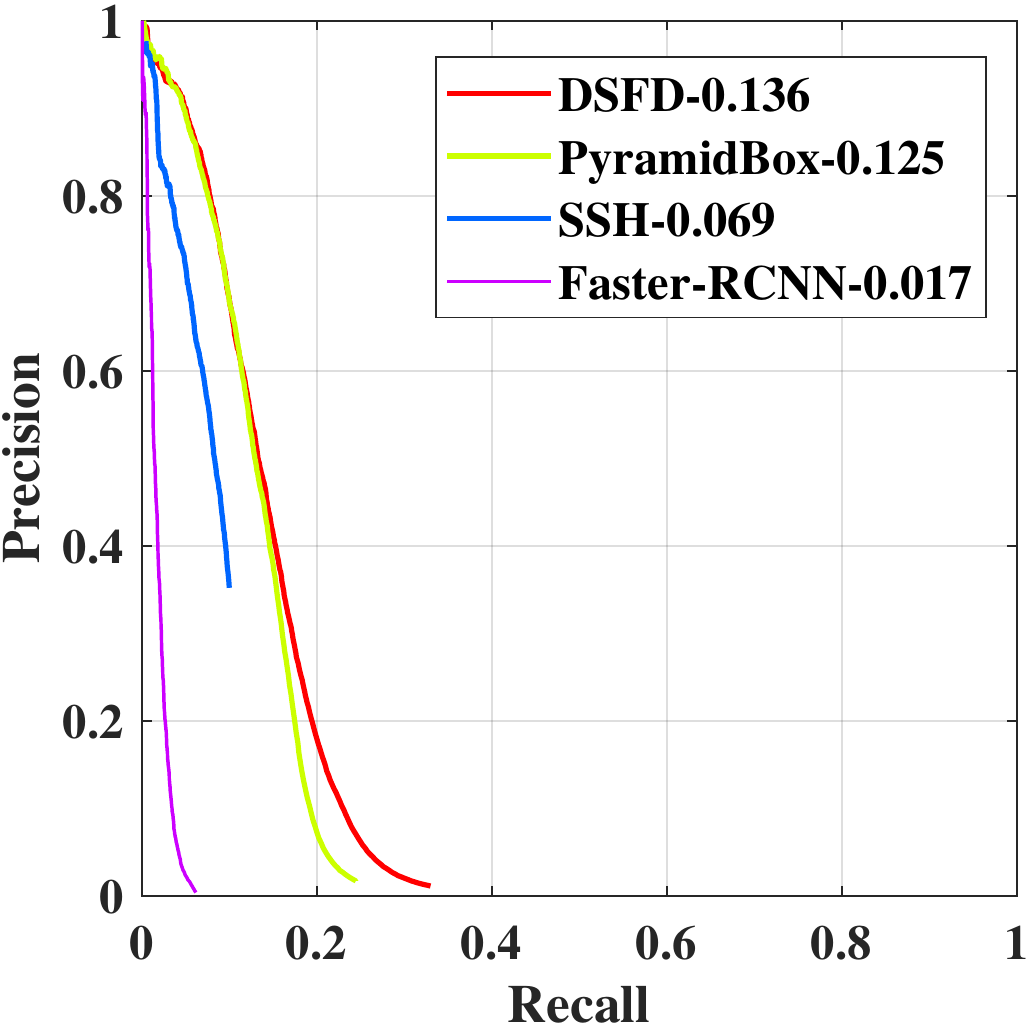}
		}\hspace{-2mm}
		\subfigure[DSFD]{
			\includegraphics[width=5.6cm]{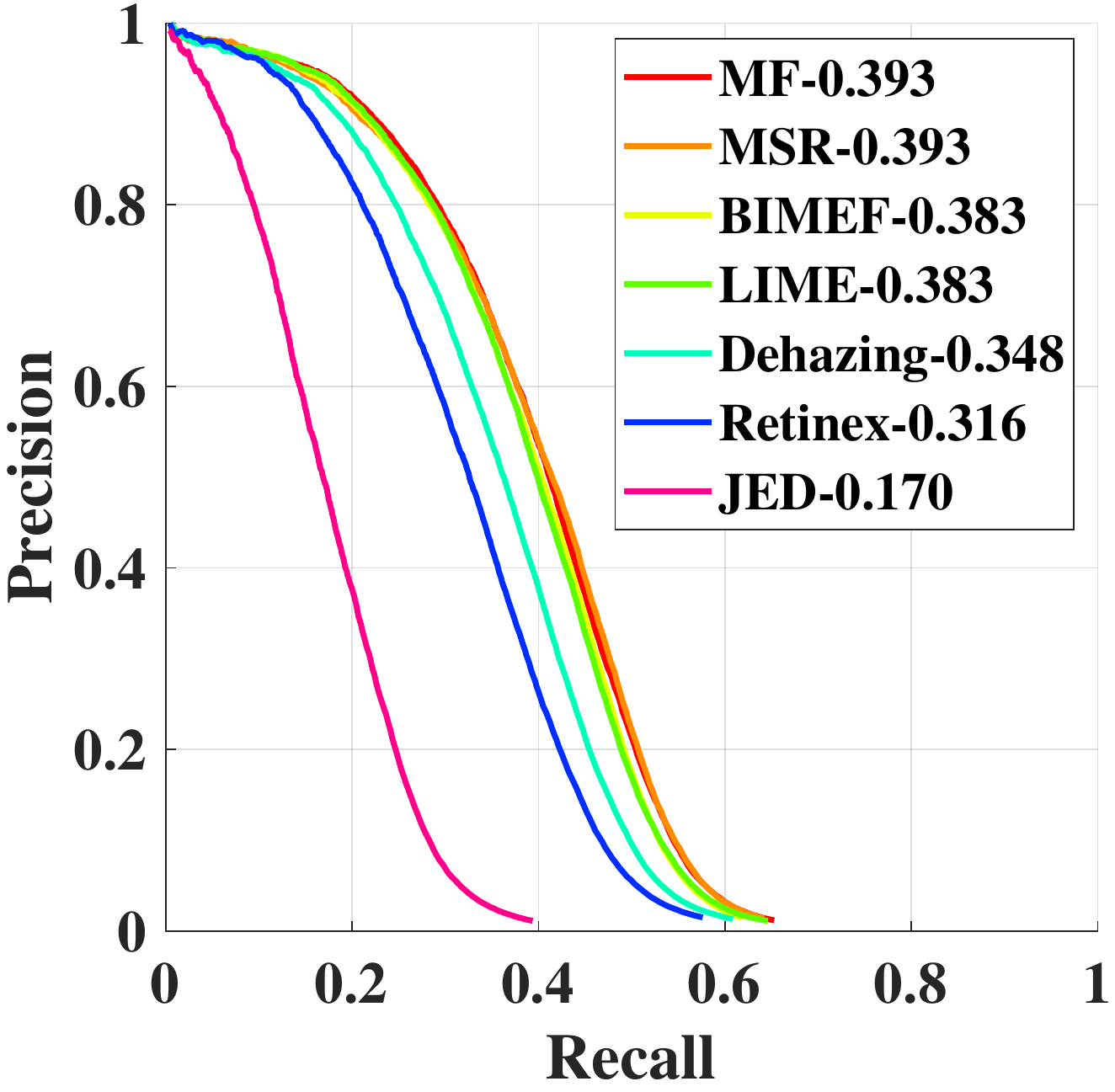}
		}\hspace{-2mm}
		\quad
		\subfigure[PyramidBox]{
			\includegraphics[width=5.6cm]{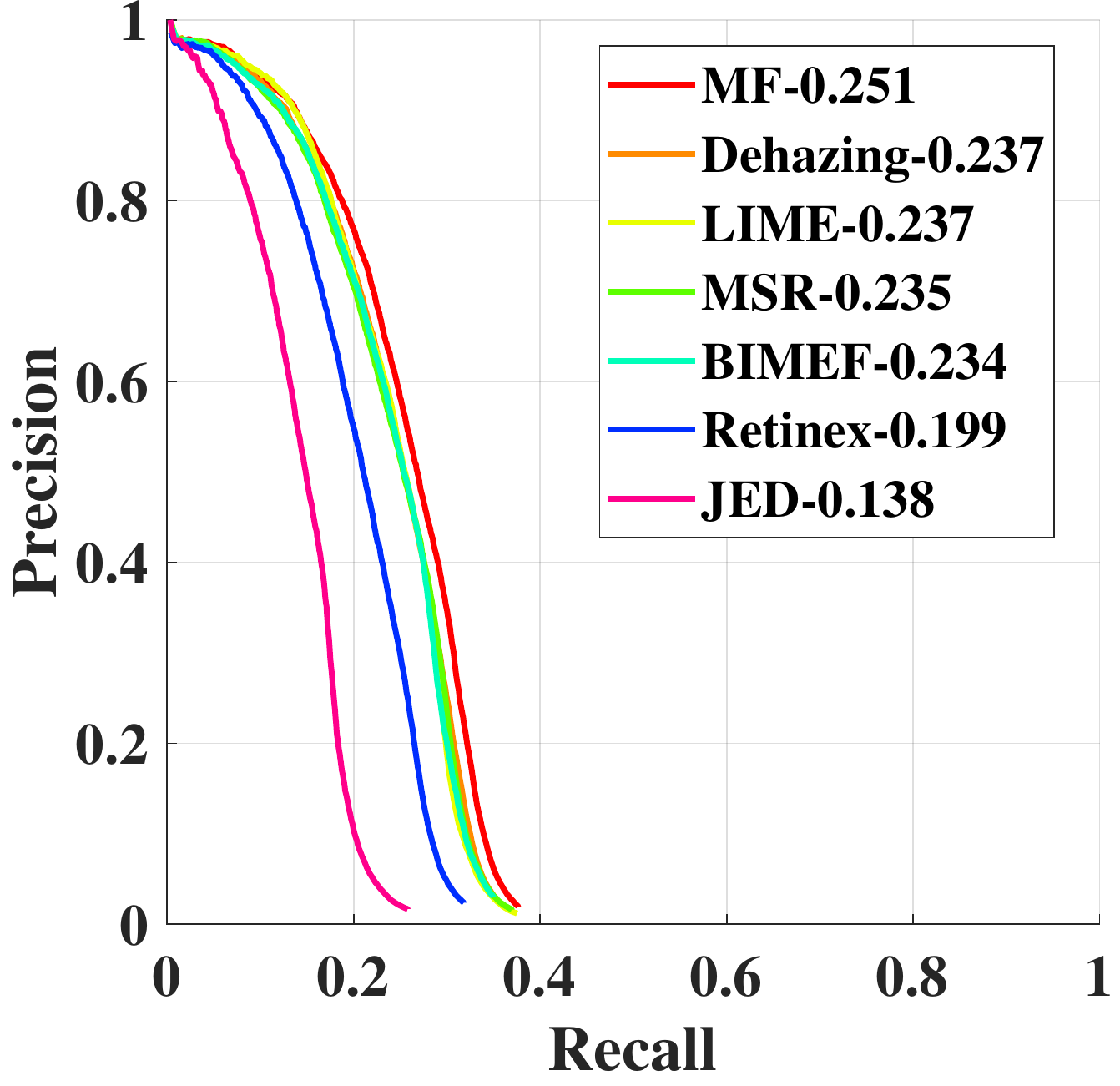}
		}
		\vspace{-2mm}
		\caption{Evaluation results of pretrained baseline on original and enhanced images of the proposed DARK FACE dataset.}
		\vspace{-7mm}
		
		\label{fig:Face_after_results}
	\end{figure*}

	We next use the enhancement algorithms to pre-process the annotated dataset and then apply the above two pre-trained face detection methods to the processed data. While the visual quality of the enhanced images is better, as expected, the detectors do perform better. As shown in Fig.~\ref{fig:Face_after_results} (b) and (c), in most instances, the precision of the detectors notably increases compared to that of the data without enhancement. 
	Various existing enhancement methods seem to result in similar improvements here. 
	
	Despite being encouraging to see, the overall performance of the detectors still drops a lot compared to normal-light datasets. The simple cascade of low light enhancement and face detectors leave much improvement room open.
	
	\subsection{Sub-challenge 2.3 Baseline Results}

	\subsubsection{Baseline Composition}
	
	We use four state-of-the-art object detection models: (1) Faster R-CNN (\textbf{FRCNN})~\cite{NIPS2015_5638}; (2) \textbf{YOLO-V3}~\cite{redmon2018yolov3}; (3) \textbf{SSD-512}~\cite{liu2016ssd}; and (4) \textbf{RetinaNet}~\cite{lin2018focal}. 	
	
	We employ five state-of-the-art deep learning-based deraining algorithms: (a) JOint Rain DEtection and Removal\footnote{http://www.icst.pku.edu.cn/struct/Projects/joint\_rain\_removal.html} (\textbf{JORDER}) \cite{yang2017deep}; (b) Deep Detail Network\footnote{\url{https://github.com/XMU-smartdsp/Removing_Rain}} (\textbf{DDN}) \cite{Fu_2017_CVPR}; (c) Conditional Generative Adversarial Network\footnote{https://github.com/TrinhQuocNguyen/Edited\_Original\_IDCGAN} (\textbf{CGAN}) \cite{zhang2017image}; (d) Density-aware Image De-raining method using a Multistream Dense Network\footnote{https://github.com/hezhangsprinter/DID-MDN} (\textbf{DID-MDN}) \cite{zhang2018density}; and (e) \textbf{DeRaindrop}\footnote{https://github.com/rui1996/DeRaindrop} \cite{qian2018attentive}. For fair comparisons, we re-train all deraining algorithms using the same provided training set.

	\subsubsection{Results and Analysis}
	Table \ref{tab-det-RID} shows mAP results comparisons for different deraining algorithms using different detection models on the held-out test set. Unfortunately, we find that almost all existing deraining algorithms deteriorate the objects detection performance compared to directly using the rainy images for YOLO-V3, SSD-512, and RetinaNet (The only exception is the detection results by FRCNN). This could be due to those deraining algorithms are not trained towards the end goal of object detection, they are unnecessary to help this goal, and the deraining process itself might have lost discriminative, semantically meaningful true information, and thus hampers the detection performance.
	In addition, Table \ref{tab-det-RID} shows that YOLO-V3 achieves the best detection performance, independent of deraining algorithms applied. We attribute this to the small objects in a relative long distance from the camera in the test set since YOLO-V3 is known to improve small object detection based on multi-scale prediction structure.

	\section{Competition Results: Overview \& Analysis}
	\label{sec:results_analysis}
	The UG$^2+$ Challenge (Track 2) in conjunction with CVPR 2019 attracted large deals of attention and participation. More than 260 teams registered; among them, 82 teams finished the dry-run and submitted their final results successfully. Eventually, 6 teams were selected as winners (including a winner and a runner-up, for each sub-challenge). 
	
	In the following, we review a part of results from those participation teams who volunteer to disclose their technical details. The full leaderboards can be found at the website~\footnote{http://www.ug2challenge.org/leaderboard19\_t2.html}.

	\subsection{Sub-challenge 2.1: Competition Results and Analysis}
	
	A total of seven teams were able to outperform our best baseline numbers (mAP 18.09). The winner and runner-up teams, \textit{HRI\_DET} and \textit{superlab403}, achieve record-high mAP results of 52.71 and 49.22, respectively. All teams used deep learning solutions. In addition to using most sophisticated networks,  several interesting observations could be concluded: 
	i) while many teams went with the dehazing-detection cascade idea (like our baselines, but usually jointly trained), the top-2 winners used end-to-end trained/adapted detection models on the hazy training set, without an (implicit) dehazing module; 
	ii) the utilization of unlabeled data seems to open up much potential, and we believe it should be paid more attention to in the future; and 
	iii) multi-scale testing and ensembling contribute to many performance gains.
	
	As the winner team, \textit{HRI\_DET} used Faster R-CNN, with the ImageNet-pretrained backbone of ResNeXt-101 \cite{xie2017aggregated} and Feature Pyramid Network (FPN) \cite{lin2017feature}. The Faster R-CNN was then tuned on the mixed dataset of MS COCO, PASCAL VOC and KITTI, with the common data augmentations. To further boost the performance, the team delved deep into the provided unlabeled hazy set, and adopted semi-supervised learning by using the unlabeled data to train the feature extractor with a reconstruction loss.  The team used a batch size of 8 and trained the network using 8 Tesla M40 GPUs for 30,000 iterations with an initial learning rate of 0.005. They also found it helpful to apply stochastic weight averaging (SWA) \cite{izmailov2018averaging} to aggregate several checkpoint models in one training pass, and further to ensemble multiple models (by averaging model weights) obtained from different training passes (\textit{e.g.} with different learning rate schedulers). During inference, a three-scale testing is performed by resizing images to $1333 \times 1000$, $1000 \times 750$ and $2100 \times 600$. The third scale is applied to a closer view of the image, by cropping the foreground region defined as the bounding box of all high-confidence predictions with the first scale.
	
	The runner-up team \textit{superlab403} chose the Cascade R-CNN \cite{cai2018cascade} baseline, and also replaced the original ResNet-101 backbone with ResNeXt-101. The team analyzed the distribution of target aspect ratio from the training set, and selected four new anchor ratios (0.8, 1.7, 2.6, 3.7) by $k$-means. The team also did a (well-appreciated) label cleaning effort. Data augmentations such as blur, illumination change and color perturbations were adopted in training. The model was trained with an SGD optimizer; the initial learning rate was set as 0.0025, then being decayed by a factor of 0.1 at epochs 8, 11, 21 and 41 (total training epoch number 50).
	
	Other teams have each developed their interesting solutions. 
	For example, the \textit{Mt. Star} team (ranked No. 3, mAP 31.24)  adopted a sequential cascade of the dehazing model (DehazeNet \cite{DehazeNet}) and the detection model (Faster-RCNN), each first pre-trained on their own and then jointly tuned end-to-end on the training set. Multi-scale testing was adopted. The \textit{ilab} team (ranked No. 6, mAP 19.15) also referred to the dehazing-detection cascade idea, but using DeblurGAN \cite{kupyn2018deblurgan} (re-trained on haze data) for the dehazing model and Yolo-V2 \cite{redmon2018yolov3} for the detection model, with a content loss. 
	
	\subsection{Sub-challenge 2.2: Competition Results and Analysis}
	
	A total of three teams were able to outperform our best baseline numbers (mAP 39.30). The winner and runner-up teams, \textit{CAS-Newcastle} and \textit{CAS\_NEU}, achieve high mAP results of 62.45 and 61.84, respectively. Similarly to Sub-challenge 2.1, all teams used deep learning solutions; yet interestingly, the most successful solutions are based on enhancement-detection cascades, showing a different trend with  Sub-challenge 2.1.
	
	The \textit{CAS-Newcastle} and \textit{CAS\_NEU} team adopted cascades of low-light enhancement (MSRCR \cite{MR}) and detection (Selective Refinement Network~\cite{Chi2019SRN} / RetinaNet~\cite{lin2018focal}) models, where the detection models were directly trained on the enhancement models' preprocessed outputs. The \textit{SCUT-CVC} team (ranked No. 7, mAP 35.18) found tone mapping \cite{drago2003adaptive} to be an impressively effective pre-processing, on top of which they tuned two DSFD detectors (with VGG-16 and ResNet-152 backbones), whose results were ensembled by late fusion.  The \textit{PHI-AI} team (ranked No. 7, mAP 29.95) adopted a U-Net \cite{ronneberger2015u} enhancer and a DSFD detector. The \textit{tjfirst} team (ranked No. 12, mAP 26.50) referred to a more sophisticated enhancement module (first enhancing illumination by LIME \cite{Guo_2017_Lime}, then super-resolving by DPSR \cite{zhang2019deep}, ended by denoising with BM3D \cite{BM3D}), followed by aggregating the DSFD-detection results on the original and enhanced images.

	\subsection{Sub-challenge 2.3: Competition Results and Analysis}
	
	Different from the first two sub-challenges, Sub-challenge 2.3 is substantially more difficult due to its ``zero-shot'' nature. Typical solutions that we see from the challenge teams include deraining + detection cascades; as well as ensembling multiple pre-trained detectors (\textit{e.g.}, the \textit{CAS-Newcastle-TUM} team). Unfortunately but not too surprisingly, none of the participation teams was able to outperform our baseline.   
	That concurs with the conclusion drawn from the recent benchmark work \cite{li2019single}: \textit{``Perhaps surprisingly at the first glance, we find that almost all existing deraining algorithms will deteriorate the detection performance compared to directly using the rainy images...''} \textit{``No existing deraining method seems to directly help detection. That may encourage the community to develop new robust algorithms to account for high-level vision problems on real-world rainy images. On the other hand, to realize the goal of robust detection in rain does not have to adopt a de-raining preprocessing; there are other domain adaptation type options...''}. 
	
\begin{figure}[t]
		\centering
		\subfigure{
			\includegraphics[width=8.5cm]{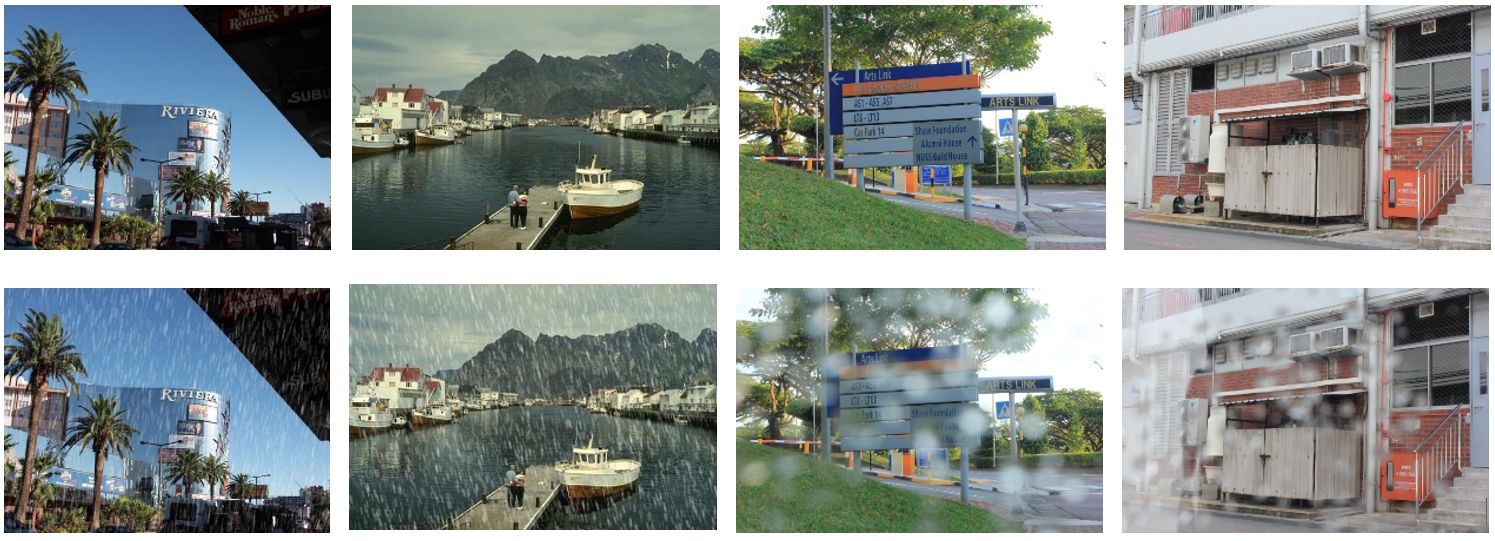}
		}
		\caption{Samples of open-source paired rain image training set only including the rain streak and raindrop-related degradation.}
		\vspace{-3mm}		
		\label{fig:subchallenge_23_1}
	\end{figure}

	In fact, our Sub-challenge 2.3 is more difficult and challenging. There is no training set that is close to the testing set provided, which fails all submitted methods. Therefore, the problem is closer to zero-shot and unsupervised learning problem.
	Existing open-source paired rain image training set, \textit{e.g.} Rain800~\cite{zhang2017image} and raindrop dataset~\cite{qian2018attentive}, usually consider only one kind of rain degradation, i.e. rain streak or raindrop. Based on a benchmark paper~\cite{li2019benchmarking}, the synthetic and captured paired images rely on three rain models. The generated rain images are not visually authentic and close to the real captured ones. Their background layers of these images are clear and the objects in these images have their appropriate sizes and locations as shown in Fig.~\ref{fig:subchallenge_23_1}.
	
	\begin{figure}[t]
		\centering
		\subfigure{
			\includegraphics[width=8.5cm]{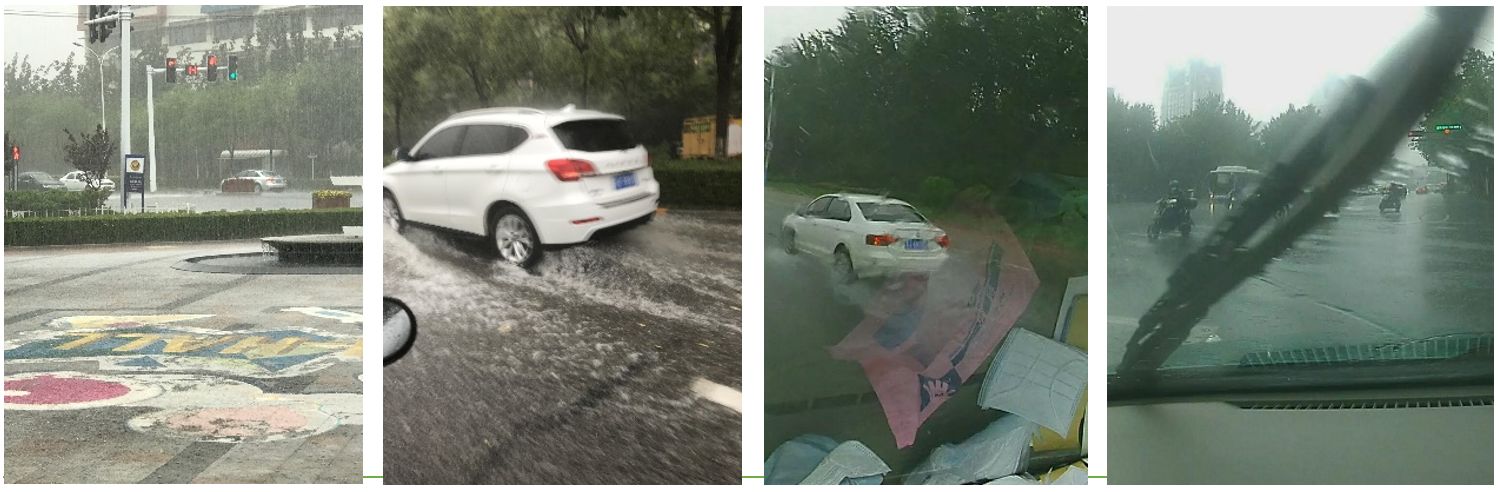}
		}
		\caption{Real rain images captured in driving or surveillance cases have degradation of rain accumulation, blurring, reflectance and occlusions.}
		\vspace{-6mm}
		\label{fig:subchallenge_23_2}
	\end{figure}
	
	The testing rain images in our Sub-challenge 2.3 are collected in real driving or surveillance scenarios. They come with other degradation such as rain accumulation, blurring, reflectance, occlusion \textit{etc}. as shown in Fig.~\ref{fig:subchallenge_23_2}, which causes the domain shift problem and makes the related restoration and detection tasks harder.
	
	Our challenging results further confirm that, the existing open-source paired datasets are poorly related when it comes to task purposes, \textit{e.g.} object detection, in real scenario. Since driving and surveillance are representative of real application scenarios where deraining may be desired, this sub-challenge is well-worth exploration. We are intended to extend this challenge to next year and look for better deraining algorithms to be proposed.
	
	\section{Discussions and Future Directions}
	\label{sec:insights}
	
	In this challenge, the submitted solutions and analysis results provide rich experiences, meaningful observations and insights, as well as potential future directions:
	\begin{itemize}
		\item Deep learning methods are preferred by all participants in their submitted solutions.
		Except that in Sub-challenge 2.2, some teams choose to apply hand-crafted low-light enhancement methods as pre-processing, other submitted methods are deep learning-based as they are flexible to be tuned to improve the performance based on the given task (dataset).
		\item 
		In Sub-challenge 2.1, the top-2 winners make efforts in exploring the potential of unlabeled data,
		via semi-supervised learning with a reconstruction loss to guide the reconstruction of the full-picture from the intermediate feature.
		This strategy leads to performance improvement, which shows that it should be paid more attention to in future.	
		\item For different tasks and techniques used to tackle the problems, the best choices of the framework might be different. In our challenge, the winner of Sub-challenge 2.1 uses a one-step detection scheme (without an implicit dehazing module) while the winner of Sub-challenge 2.2 takes the cascade of enhancement and detection.
		\item Some teams, \textit{e.g.} \textit{tjfirst}, report the effectiveness to apply a sequential enhancement process to remove mixed degradation, such as under-exposure, low-resolution blurring, and noise.
		It also shows a valuable path that is worth further exploration.
		\item Separate consideration of enhancement and detection might lead to deteriorated performance in detection.
		In Table~\ref{dehaze_res}, the offline dahazing operation is first applied by AOD-Net, DCPDN, MSCNN. The results are taken as the input of object detection methods, \textit{i.e.} RetinaNet, Mask R-CNN, YOLO-V3, FPN. Many combination groups of dehazing operation and object detection method obtain inferior results than directly applying object detection without any pre-enhancement, such as (AOD-Net, RetinaNet) for person category (54.93 $<$ 55.85) and (AOD-Net, Mask R-CNN), (AOD-Net, Mask R-CNN), (AOD-Net, Mask R-CNN) for all categories (2.81,2.83,2.99 $<$ 3.37). In Table~\ref{tab-det-RID}, YOLO-V3, SSD-512 and RetinaNet generate worse object detection results if pre-deraining is applied.
		\item 
		The existing results show the difficulty of three tasks we set.
		In Sub-challenge 2.1 and 2.2, the winners only achieve the results below 65 MAP, much inferior to the performance on datasets with degradation.
		In Sub-challenge 2.3, no participates achieve the results superior to the baseline results.
		The results show that, our datasets are challenging and there is still large room for further improvement.
	\end{itemize}
	
	\section{Conclusions}
	\label{sec:conclusion}
	
	As concurred by most teams in the post-challenge feedbacks, it is widely agreed that the three sub-challenges in he UG$^{2+}$ challenge 2019 Track 2 represent a very difficult, under-explored, yet high meaningful class of computer vision problems in practice. While some promising progress has been witnessed from the large volume of team participation~\footnote{
		Taiheng Zhang is with the Department of Mechanical Engineering, Zhejiang University, Hangzhou 310027, China (email: thzhang@zju.edu.cn).
		
		Qiaoyong Zhong, Di Xie and Shiliang Pu are with the Hikvision Research Institute, Hangzhou 310051, China (e-mail: zhongqiaoyong@hikvision.com; xiedi@hikvision.com; pushiliang.hri@hikvision.com).
		
		
		Hao Jiang, Siyuan Yang, Yan Liu, Xiaochao Qu, Pengfei Wan are with the Mtlab, Meitu Inc., Beijing 100080, China (e-mail: jh1@meitu.com, ysy2@meitu.com, ly33@meitu.com, qxc@meitu.com, wpf@meitu.com).
		
		Shuai Zheng, Minhui Zhong, Lingzhi He, Zhenfeng Zhu and Yao Zhao are with the Insitute of Information Science, Beijing Jiaotong University, Beijing, 100044, China (e-mail: ericzheng1997@163.com, zylayee@163.com, 19112002@bjtu.edu.cn, zhfzhu@bjtu.edu.cn, yzhao@bjtu.edu.cn).

		Taiyi Su is with the Department of Computer Science and Technology, Tongji University, Shanghai, 201804, China (e-mail: tysu@tongji.edu.cn).
		Yandong Guo is with XPENGMOTORS, Beijing, China (e-mail: guoyd@xiaopeng.com).
		
		
		Jinxiu Liang, Tianyi Chen, Yuhui Quan and Yong Xu are with the School of Computer Science and Engineering at South China University of Technology, Guangzhou 510006, China (e-mail: cssherryliang@mail.scut.edu.cn; csttychen@mail.scut.edu.cn; csyhquan@scut.edu.cn; yxu@scut.edu.cn).
		
		Jingwen Wang is with Tencent AI Lab, Shenzhen 518000, China (e-mail: jwongwang@tencent.com).
		
		Shifeng Zhang, Chubing Zhuang and Zhen Lei are with the Institute of Automation of the Chinese Academy of Sciences, Beijing 100190, China (e-mail: shifeng.zhang@nlpr.ia.ac.cn; chubin.zhuang@nlpr.ia.ac.cn; zlei@nlpr.ia.ac.cn).
		
		Jianning Chi, Huan Wang and Yixiu Liu are with the Northeastern University, Shenyang, 110819, China (e-mail: chijianning@mail.neu.edu.cn; 1870692@stu.neu.edu.cn; yixiu9713@foxmail.com).
		
		Huang Jing, Guo Heng and Jianfei Yang are with the Nanyang Technological University, 639798, Singapore (e-mail: jhuang027@e.ntu.edu.sg;  hguo007@e.ntu.edu.sg; yang0478@e.ntu.edu.sg).
		
		Zhenyu Chen is with the Big Data Center, State Grid Corporation of China, Beijing, China and the China Electric Power Research Institute, Beijing 100031, China (e-mail: czy9907@gmail.com).
		
		Dong Yi is with the Winsense Inc., Beijing 100080, China (e-mail: yidong@winsense.ai).
		
		Cheng Chi, Kai Wang, Jiangang Yang, Likun Qin are with the University of Chinese Academy of Sciences, Beijing 100049, China (chicheng15@mails.ucas.ac.cn; wk2008cumt@163.com; yangjiangang18@mails.ucas.ac.cn; qinlikun19@ucas.ac.cn).
		
		Liguo Zhou and Mingyue Feng are with the Department of Informatics, Technical University of Munich, Garching 85748, Germany (e-mail: liguo.zhou@tum.de; mingyue.feng@tum.de).
		
		Chang Guo, Yongzhou Li and Huicai Zhong are with the Institute of Microelectronics of the Chinese Academy of Sciences, Beijing 100190, China (e-mail: guochang@ime.ac.cn; liyongzhou@ime.ac.cn; zhonghuicai@ime.ac.cn).
		
		Xingyu Gao is with the Institute of Microelectronics of the Chinese Academy of Sciences, Beijing, China and the Beijing Jiaotong University, Beijing 100044, China (e-mail: gaoxingyu@ime.ac.cn).
		
		Stan Z. Li is with the Institute of Automation of the Chinese Academy of Sciences, Beijing and the Westlake University, Hangzhou 310024, China (e-mail: szli@nlpr.ia.ac.cn).
		
		Zheming Zuo is with the Department of Computer Science, Durham University, Durham, United Kingdom (e-mail: zheming.zuo@durham.ac.uk).
		
		Wenjuan Liao is with the Australian National University, Acton ACT 0200, Australia (e-mail: wenjuan.liao@anu.edu.au).
		
		Ruizhe Liu is with the Sunway-AI Co., Ltd, Zhuhai, China and the Chinese Academy of Sciences R\&D Center for Internet of Things, Wuxi 214200, China (e-mail: liuruizhe@ciotc.org).
		
		Shizheng Wang is with the Institute of Microelectronics of Chinese Academy of Sciences, Beijing, China and the Chinese Academy of Sciences R\&D Center for Internet of Things, Wuxi 214200, China (e-mail: shizheng.wang@foxmail.com).
		
	}, there remains large room to be improved. Through organizing this challenge, we expect to evoke a broader attention from the research community to address these challenges, which are barely covered by previous benchmarks. We look forward to making UG$^{2+}$ a recurring event and also evolving/updating our problems and datasets every year.
	
	\begin{table*}[!ht]\scriptsize
		\caption{Detection results (mAP) on the held-out test set.}
		\vspace{-2mm}
		\begin{center}\small{
				\begin{tabular}{c|c|c|c|c|c|c}
					\hline
					& Rainy  & JORDER \cite{yang2017deep}  & DDN \cite{Fu_2017_CVPR} & CGAN \cite{zhang2017image} & DID-MDN \cite{zhang2018density}  & DeRaindrop \cite{qian2018attentive} \\
					\hline
					FRCNN~\cite{NIPS2015_5638}        & 16.52   & 16.97  & 18.36 & \textbf{23.42}  & 16.11  & 15.58  \\
					YOLO-V3~\cite{redmon2018yolov3} & \textbf{27.84}   & 26.72  & 26.20  & 23.75  & 24.62  & 24.96  \\
					SSD-512~\cite{liu2016ssd}         & \textbf{17.71}   & 17.06  & 16.93  & 16.71  & 16.70  & 16.69 \\
					RetinaNet~\cite{lin2018focal}     & \textbf{23.92}   & 21.71  & 21.60  & 19.28  & 20.08  & 19.73  \\            
					\hline
			\end{tabular}}
			\vspace{-4mm}
			\label{tab-det-RID}
		\end{center}
	\end{table*}
	
	\begin{figure}[t]
		\centering
		\subfigure{
			\includegraphics[width=4cm]{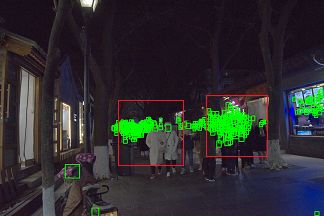}
		} \hspace{-1.5mm}
		\subfigure{
			\includegraphics[width=4cm]{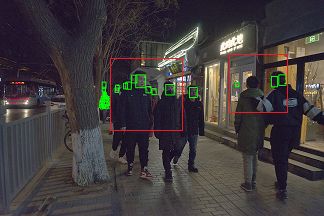}
		} 
		\\
		\vspace{-2mm}
		\subfigure{
			\includegraphics[width=1.9cm]{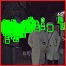}
		}\hspace{-1.5mm}
		\subfigure{
			\includegraphics[width=1.9cm]{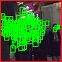}
		}
		\hspace{-1.5mm}
		\subfigure{
			\includegraphics[width=1.9cm]{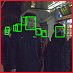}
		}\hspace{-1.5mm}
		\subfigure{
			\includegraphics[width=1.9cm]{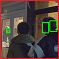}
		}	
		\\
		\vspace{-2mm}
		\caption{Sample face detection results of pretrained baseline on the enhanced images of the proposed DARK FACE dataset. \wh{The face regions in the red boxes are zoomed-in for better viewing.}}
		\vspace{-7mm}		
		\label{fig:Face_after_results_sub}
	\end{figure}
	
	\bibliographystyle{IEEEtran}
	\scriptsize
	

\end{document}